\documentclass[10pt,twocolumn,letterpaper]{article}
\usepackage{cvpr}              

\usepackage{graphicx}
\usepackage{amsmath}
\usepackage{amssymb}
\usepackage{amsthm}
\usepackage{booktabs}
\usepackage[table]{xcolor}
\usepackage[pagebackref,breaklinks,colorlinks]{hyperref}
\usepackage[capitalize]{cleveref}
\crefname{section}{Sec.}{Secs.}
\Crefname{section}{Section}{Sections}
\Crefname{table}{Table}{Tables}
\crefname{table}{Tab.}{Tabs.}

\newtheorem{property}{Property}

\def\cvprPaperID{410}
\def\confName{CVPR}
\def\confYear{2022}

\begin{document}

\title{Attributable Visual Similarity Learning}

\author{Borui Zhang, Wenzhao Zheng, Jie Zhou, Jiwen Lu\thanks{Corresponding author.}\\\
{Department of Automation, Tsinghua University, China}\\
{Beijing National Research Center for Information Science and Technology, China}\\
{\tt\small \{zhang-br21, zhengwz18\}@mails.tsinghua.edu.cn; \{jzhou, lujiwen\}@tsinghua.edu.cn}
}
\maketitle

\begin{abstract}
This paper proposes an attributable visual similarity learning (AVSL) framework for a more accurate and explainable similarity measure between images. 
Most existing similarity learning methods exacerbate the unexplainability by mapping each sample to a single point in the embedding space with a distance metric (e.g., Mahalanobis distance, Euclidean distance).
Motivated by the human semantic similarity cognition, we propose a generalized similarity learning paradigm to represent the similarity between two images with a graph and then infer the overall similarity accordingly.
Furthermore, we establish a bottom-up similarity construction and top-down similarity inference framework to infer the similarity based on semantic hierarchy consistency.
We first identify unreliable higher-level similarity nodes and then correct them using the most coherent adjacent lower-level similarity nodes, which simultaneously preserve traces for similarity attribution.
Extensive experiments on the CUB-200-2011, Cars196, and Stanford Online Products datasets demonstrate significant improvements over existing deep similarity learning methods and verify the interpretability of our framework. \footnote{Code: \url{https://github.com/zbr17/AVSL}.}
\end{abstract}
\vspace{-5mm}
\section{Introduction} \label{sec:Introduction}

Similarity learning is a fundamental task in the field of computer vision, where most prevalent works (i.e., metric learning methods) employ a distance metric to measure the similarities between samples. 
They transform features into an embedding space and define the dissimilarity as the Euclidean distance in this space, where the objective is to cluster similar samples together and separate dissimilar ones apart from each other. 
While conventional methods use hand-crafted features like SIFT~\cite{lowe2004distinctive} and LBP~\cite{ahonen2006face}, deep metric learning methods employ convolutional neural networks (CNNs)~\cite{krizhevsky2012imagenet} to extract more representative features and demonstrate superior performance.
In recent years, similarity learning has been widely applied to various vision tasks such as face recognition~\cite{hu2014discriminative, taigman2014deepface, schroff2015facenet}, person re-identification~\cite{chen2017beyond, hermans2017defense, lin2017consistent, yu2018hard}, and image classification~\cite{lu2015multi, chen2019hybrid}.

\begin{figure}[tb]
    \centering
    \includegraphics[width=1\linewidth]{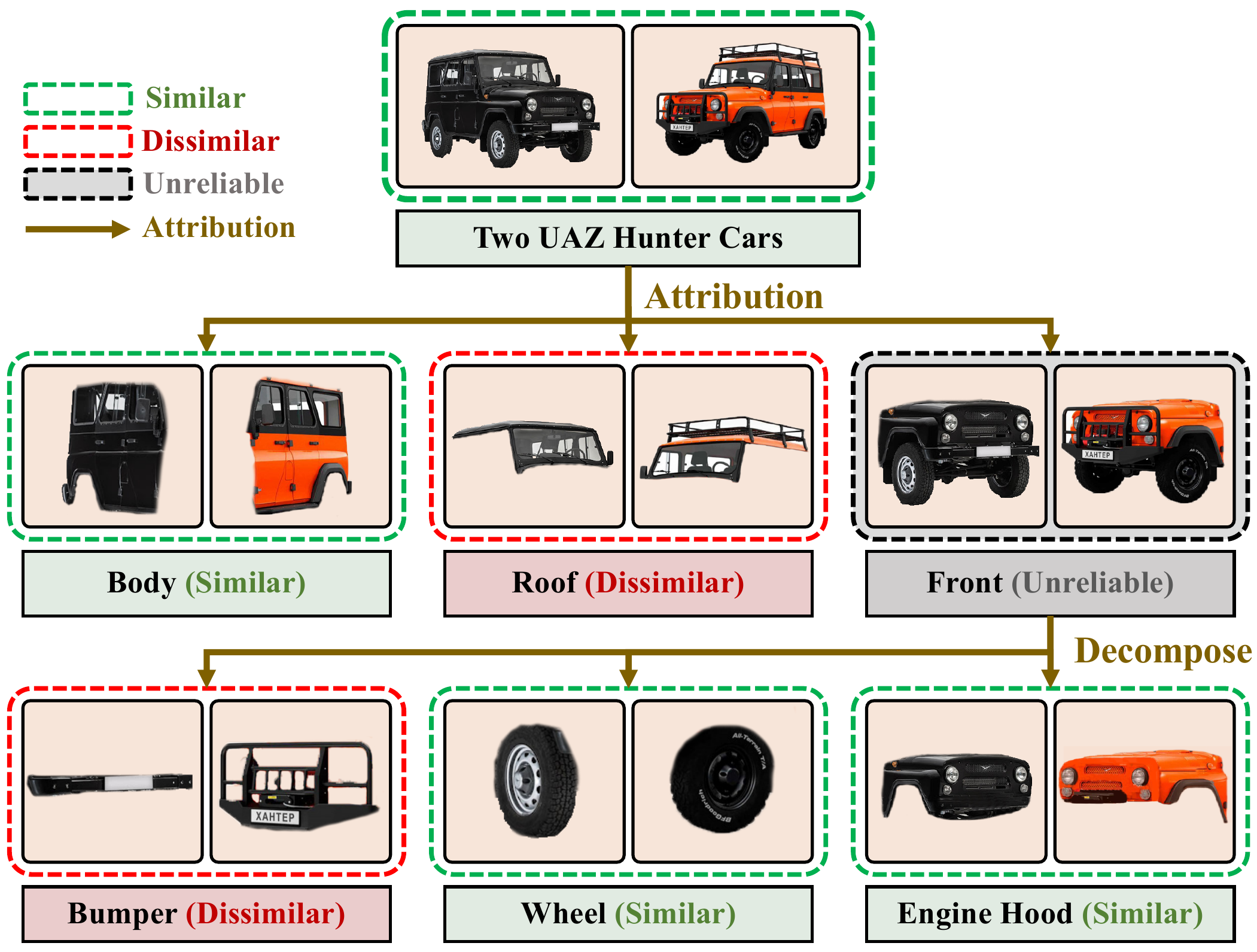}
    \vspace{-6mm}
    \caption{The motivation of the proposed AVSL framework. 
    Humans recognize each image as a complex set of concepts and compare two images hierarchically~\cite{lake2015human}. 
    For example, when inferring the similarity between two cars, humans usually first compare high-level features such as shapes or colors and then turn to finer features such as wheel structures when a coarse observation does not distinctly distinguish them. 
    Motivated by this, we propose to employ a graph structure to decompose sample pairs into discriminative concept nodes, which is more consistent with how humans perceive the cognitive distance and is beneficial to the attribution of the similarity measurement.}
    \label{fig:motivation}
    \vspace{-6mm}
\end{figure}

The essential goal of visual similarity learning is to obtain a similarity measure that generalizes well to unseen data.
It has been shown that the good generalization of the human visual system comes from the ability to parse objects into parts and relations and learn the underlying concepts~\cite{lake2015human}.
Humans also infer the similarity between two images hierarchically by first comparing high-level features and then delving into lower-level features, as illustrated in Figure~\ref{fig:motivation}.
However, most existing similarity learning methods simply project each sample to one single vector and employ the Mahalanobis distance or Euclidean distance as the similarity function.
They only use the top-level feature to represent an image and directly compute the similarity without inference. 
Also, using a single vector for similarity measure exacerbates the unexplainability caused by the black-box CNNs and leads to untraceable similarity measurement, i.e., we can hardly attribute the overall similarity to specific features.  
To alleviate this issue, some methods~\cite{stylianou2019visualizing, chen2020adapting, zhu2019visual} attempt to extend neural network visualization techniques to deep metric learning and generate a saliency map for each image. 
Still, they treat the similarity computing model as a black box and can only explain it subjectively in a post hoc way, where the similarity computing process remains untraceable and unexplainable.

In this paper, we propose an attributable visual similarity learning (AVSL) framework to actively explain the learned similarity measurement. 
We generalize the prevalent metric learning paradigm to represent the similarity between images by a graph and then analyze it to infer the overall similarity. 
We use CNNs to extract hierarchical visual features in a bottom-up manner, where higher-level features encode more abstract concepts~\cite{zeiler2014visualizing, yosinski2014transferable} and can be regarded as a combination of low-level features~\cite{zeiler2010deconvolutional,zhang2018interpreting}. 
We further construct an undirected graph to represent the similarity between images.
We then propose a top-down similarity inference method based on hierarchy consistency.
We start from high-level similarity nodes and rectify identified unreliable nodes using adjacent low-level similarity nodes until reaching the lowest level, which is similar to how humans compare two objects from coarse to fine. 
The overall similarity can be easily attributed to the effect of each similarity node corresponding to certain visual concepts. 
Our framework can be readily applied to existing deep metric learning methods with various loss functions and sampling strategies. 
Extensive experiments on the widely used CUB-200-2011~\cite{wah2011caltech}, Cars196~\cite{krause20133d}, and Stanford Online Products~\cite{oh2016deep} datasets demonstrate that our AVSL framework can significantly boost the performance of various deep metric learning methods and achieve state-of-the-art results.
We also conduct visualization experiments to demonstrate the attributability and interpretability of our method.

\section{Related Work}

\paragraph{Similarity learning:}
Similarity learning aims to learn a similarity function to accurately measure the semantic similarities between images.
Conventional methods adopt the Mahalanobis distance to learn linear metric functions and further use kernel tricks to model nonlinear relations. 
Recent deep metric learning methods employ convolutional neural networks to learn an embedding space and use the Euclidean distance for similarity measurement, where the majority of works focus on designing different loss functions~\cite{hadsell2006dimensionality, schroff2015facenet, sohn2016improved, wang2019multi, sun2020circle, movshovitz2017no, qian2019softtriple, kim2020proxy} and sampling strategies~\cite{schroff2015facenet, harwood2017smart, yuan2017hard, zheng2019hardness, wu2017sampling, opitz2018deep, duan2019deep, xu2019deep, duan2018deep} for more effective training of the metric. 
For example, the contrastive loss~\cite{hadsell2006dimensionality} pulls positive pairs together while pushing negative ones farther than a fixed margin.
Song~et al.~\cite{oh2016deep} further proposed a lifted structured loss considering the global connections among a mini-batch.
Movshovitz~et al.~\cite{movshovitz2017no} simplified the pair sampling to linear complexity by including proxies in the loss formulations.
Still, an appropriate sampling strategy has been proven to be effective to boost performance.
For example, Schroff~et al.~\cite{schroff2015facenet} presented a semi-hard sampling strategy to select informative samples while discarding outliers.
Harwood~et al.~\cite{harwood2017smart} proposed a smart sampling strategy adaptive to different training stages.

Other works explore different designs of the similarity function to improve the performance.
For example, Yuan~et al.~\cite{yuan2019signal} and Huang~et al.~\cite{huang2016local} proposed an SNR distance and a PDDM module, respectively, to better guide the training process but still uses the conventional Euclidean distance during testing. 
Verma~et al.~\cite{verma2012learning} learned hierarchical distance metrics based on the class taxonomy.
Ye~et al.~\cite{ye2016makes} employed a set of metrics to describe similarities from different perspectives.
However, all the aforementioned methods represent dissimilarity by projecting samples into single points in the Euclidean distance which implies the triangle inequation,
while the proposed AVSL framework represents samples in a graph manner to model relations between concepts.
Zheng~et al.~\cite{zheng2021deep2} also exploited relations by projecting samples with multiply embedders to learn a sub-space structure. 
Differently, we propose to decompose the overall similarity hierarchically with hierarchy consistency as the inductive bias and employ a top-down similarity structure compatible with bottom-up similarity construction.
\vspace{-3mm}

\paragraph{Explainable artificial intelligence:}
Explainable artificial intelligence (XAI) has attracted considerable attention in recent years, resulting from the demand for stabler and safer AI applications. 
One category of works aims to interpret the outputs of black-box models by visualization or imitation~\cite{ribeiro2016should, zhang2018interpreting, stylianou2019visualizing, zhu2019visual, zeiler2014visualizing, zhou2016learning, chen2020adapting, yosinski2015understanding, selvaraju2017grad} (i.e., passive methods).
For example, Zeiler~et al.~\cite{zeiler2014visualizing} and Selvaraju~et al.~\cite{selvaraju2017grad} projected hidden feature maps into the input space using deconvolution and gradients, respectively, which can assist humans to understand the semantics of the hidden layers.
Ribeiro~et al.~\cite{ribeiro2016should} and Zhang~et al.~\cite{zhang2018interpreting} employed linear regressions and graph models to imitate complex rules in the black-box inference process.
Another category of works attempts to modify the model architecture to improve its explainability~\cite{wu2018beyond, zhang2018interpretable, wan2020nbdt} (i.e., active methods). 
For instance, Zhang~et al.~\cite{zhang2018interpretable} restricted each kernel of hidden layers to encoding a single concept.
Wu~et al.~\cite{wu2018beyond} proposed a tree regularization loss to favor models that can be more easily approximated by a simple decision tree.

A few works~\cite{stylianou2019visualizing, chen2020adapting, zhu2019visual, zhao2021towards} seek to extend neural network visualization techniques for deep metric learning. 
Nevertheless, they can only obtain global saliency maps and can hardly conduct quantitive attribution analysis of overall similarities, which cannot provide detailed interpretations of similarity models.
To the best of our knowledge, we are the first to explore an attributable and explainable similarity learning framework. Imitating humans to compare objects from coarse to fine, the proposed AVSL can attribute overall similarities to hierarchical hidden concepts.

\section{Proposed Approach}

In this section, we first present a generalized similarity learning paradigm and then elaborate on the proposed bottom-up similarity construction and top-down similarity inference.
Finally, we present the AVSL framework and demonstrate how to quantitatively attribute the similarity to different levels of features under our framework. 

\subsection{Generalized Similarity Learning Paradigm} \label{sec:formulation}

Let $\mathbf{X} = \{\mathbf{x}^{(1)}, \mathbf{x}^{(2)}, \cdots, \mathbf{x^{(N)}}\}$ denotes the image set, 
where sample $\mathbf{x}^{(n)} \in \mathbf{X}$ has a label $l^{(n)} \in \{l_1,l_2,\cdots,l_C\}$ with $C$ being the number of classes. 
Given an $L$-layer CNN $f$ and a sample $\mathbf{x}$, we call the $l$-layer outputs as feature maps, denoted as $\mathbf{z}^l = f^l(\mathbf{x}) \in \mathbb{R}^{c_l \times h_l \times w_l}$,
where $c_l,h_l$, and $w_l$ denote channel, height, and weight respectively.
Then a pooling operation $g^l(\cdot)$ reduces feature maps to vectors $\mathbf{v}^l = g^l(\mathbf{z}^l) \in \mathbb{R}^{c_l}$.
Existing deep metric learning methods usually add a linear projector $h^l(\cdot)$ to map $\mathbf{v}^l$ into an $r$-dimension embedding space: $\mathbf{e}^l = h^l(\mathbf{v}^l) \in \mathbb{R}^r$,
where the dissimilarity between two images $\mathbf{x}, \mathbf{x}'$ is 
$ \hat{d}(\mathbf{x}, \mathbf{x}')
= d(\mathbf{e}^l, \mathbf{e}'^l)
= \Vert \mathbf{e}^l - \mathbf{e}'^l \Vert_2
$.
For simplicity, We use similarity and dissimilarity interchangeably unless stated otherwise.

However, existing deep similarity learning methods only utilize the features from the top layer while discarding those from the hidden layers, which might contain complementary information.
To address this, we propose a generalized similarity learning (GSL) paradigm, which constructs an undirected graph $\mathcal{H}$ involving embeddings from each layer to compute overall similarities.
We denote each element of the embedding $\mathbf{e}^l$ as $e^l_i$, and define the \textbf{similarity node} of $\mathcal{H}$ as $\delta^l_i = \vert e^l_i - e'^l_i \vert$. 
In addition, the edge $\omega_{ij}$ of $\mathcal{H}$ will be elaborated in Section~\ref{sec:construction}.
To sum up, the GSL paradigm is composed of two modules:
\begin{itemize}
    \item \textbf{Similarity construction}: compute similarity nodes $\delta^l_i$ and edges $\omega_{ij}$ to construct an undirected graph $\mathcal{H}$.
    \item \textbf{Similarity inference}: infer the overall similarity $d$ according to the graph $\mathcal{H}$.
\end{itemize}

\begin{figure}
    \centering
    \includegraphics[width=0.45\textwidth]{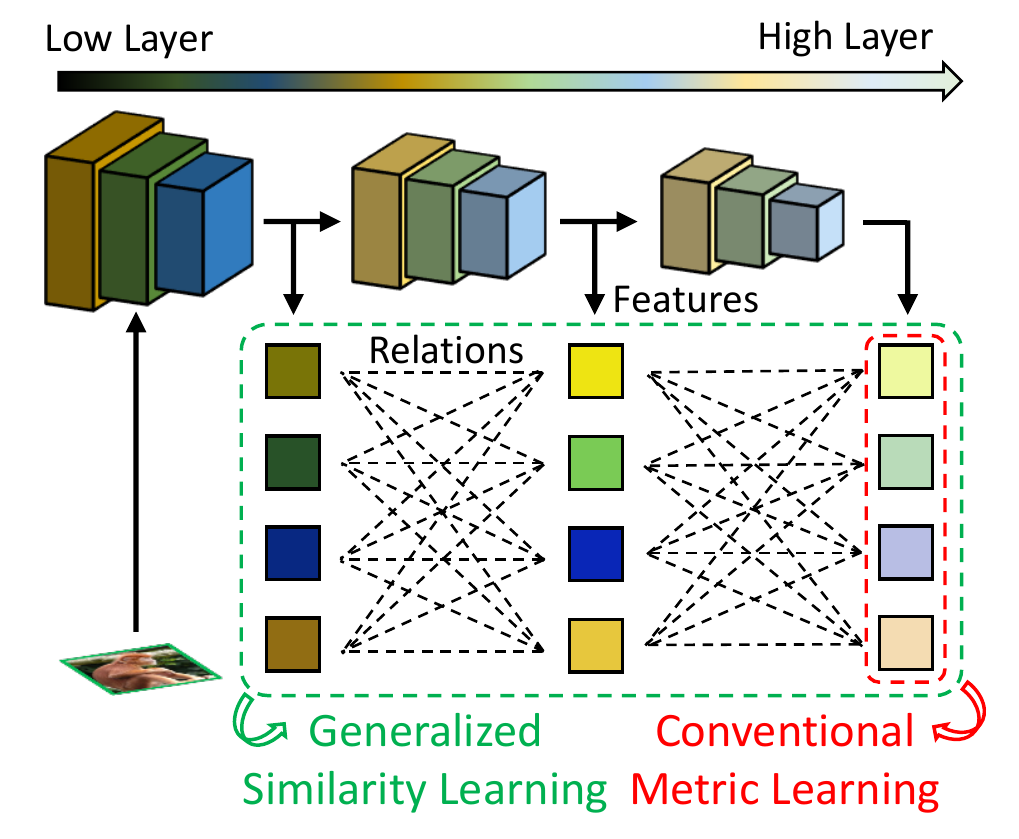}
    \caption{An illustration to show the difference between the proposed GSL and conventional DML. After extracting hierarchical features by applying CNNs, the conventional metric learning methods mainly focus on top-level features while the proposed GSL takes full advantage of features from all layers and the interactions between them to construct the similarity graph.}
    \label{fig:GSL_paradigm}
    \vspace{-5mm}
\end{figure}

The conventional metric learning methods can be regarded as a special case of GSL paradigm as shown in Figure~\ref{fig:GSL_paradigm}
when only constructing $\mathcal{H}$ with the top layer similarity nodes $\delta^L_i$ and defining the overall similarity as 
$d = \sum_{i=1}^r (\delta^L_i)^2$.

Taking full advantage of the hierarchy consistency in deep CNNs, we further propose an attributable visual similarity learning (AVSL) framework composed of bottom-up similarity construction and top-down similarity inference.

\subsection{Bottom-Up Similarity Construction} \label{sec:construction}

Conventional similarity learning methods only use embeddings from the top layer to compute the overall similarity, making it difficult to trace back to different concepts, which are encoded by embeddings from all layers as evidenced by Zhang~et al.~\cite{zhang2018interpreting}.
Different levels of features encode different levels of concepts containing complementary information, where the large receptive fields of high-level features enable them to represent high-level semantic information and omit some high-frequency details, while low-level features can capture detailed information such as textures but fail to perceive global semantics due to the restricted receptive fields. 
Still, high-level features can be regarded as the combination of low-level features ~\cite{zhang2018interpreting, zeiler2010deconvolutional}, and their connections can be exploited for the subsequent similarity inference.
Therefore, we propose a bottom-up similarity construction method to compute different levels of similarity nodes and the connections between them. 

The first step is to compute similarity nodes $\delta_i^l$. 
We extract the feature map $\mathbf{z}^l$ of the $l$-th layer by convolutional blocks from bottom to top,
and then obtain the feature vector $\mathbf{v}^l$ using global pooling.
Subsequently, we employ a fully connected layer to map the feature vector to the corresponding embedding $\mathbf{e}^l$.
Finally we obtain the similarity nodes by computing the square of the difference between normalized embeddings $\mathbf{\tilde{e}}^l, \mathbf{\tilde{e}}'^l$:
\begin{eqnarray} \label{equ:nodes}
    \delta^l_i = \vert \tilde{e}^l_i - \tilde{e}'^l_i \vert^2.
\end{eqnarray}

\begin{figure}[tb]
    \centering
    \begin{subfigure}{0.48\linewidth}
        \includegraphics[width=1\linewidth]{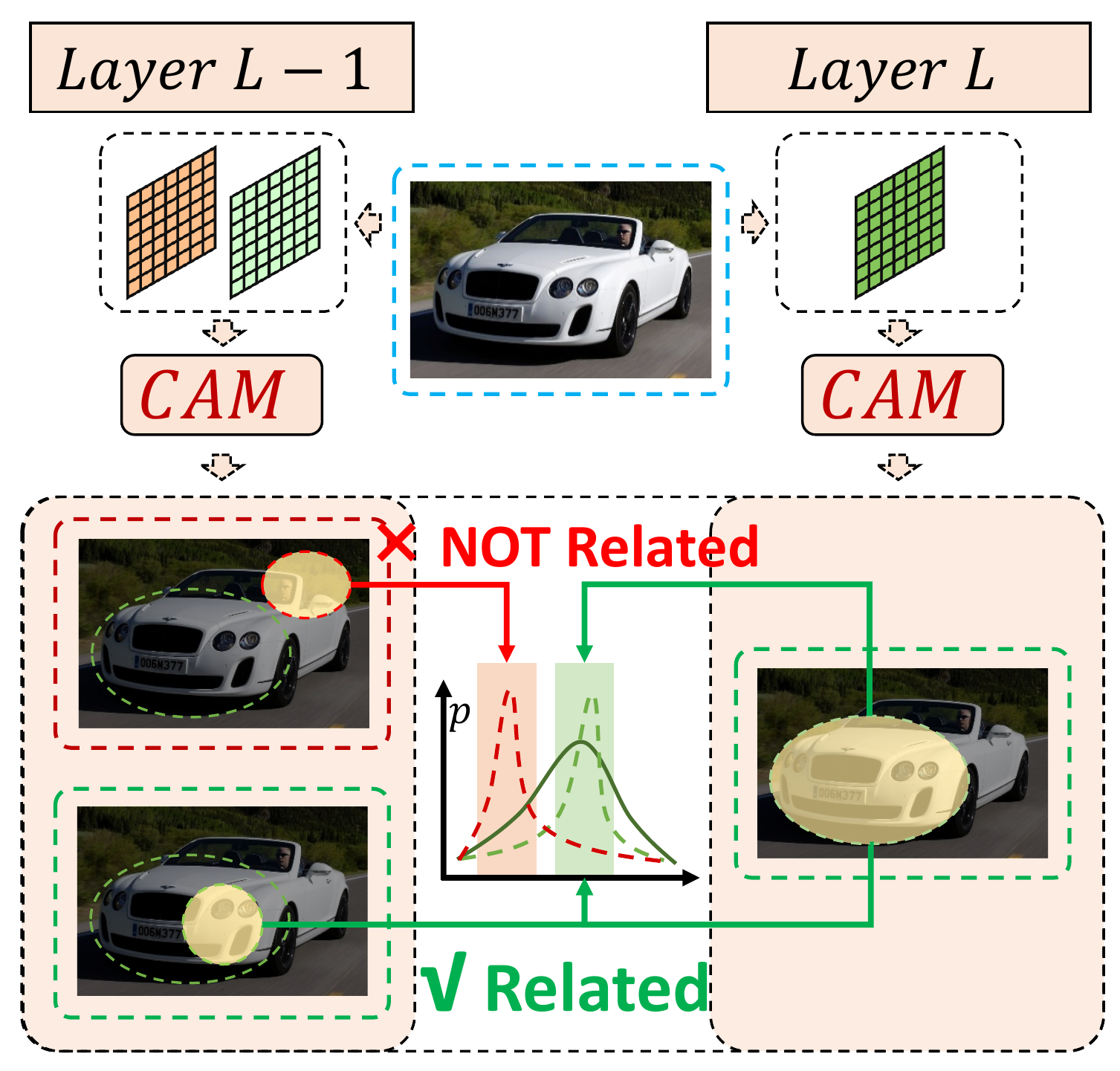}
        \caption{}
        \label{fig:cam}
    \end{subfigure}
    \begin{subfigure}{0.51\linewidth}
        \includegraphics[width=1\linewidth]{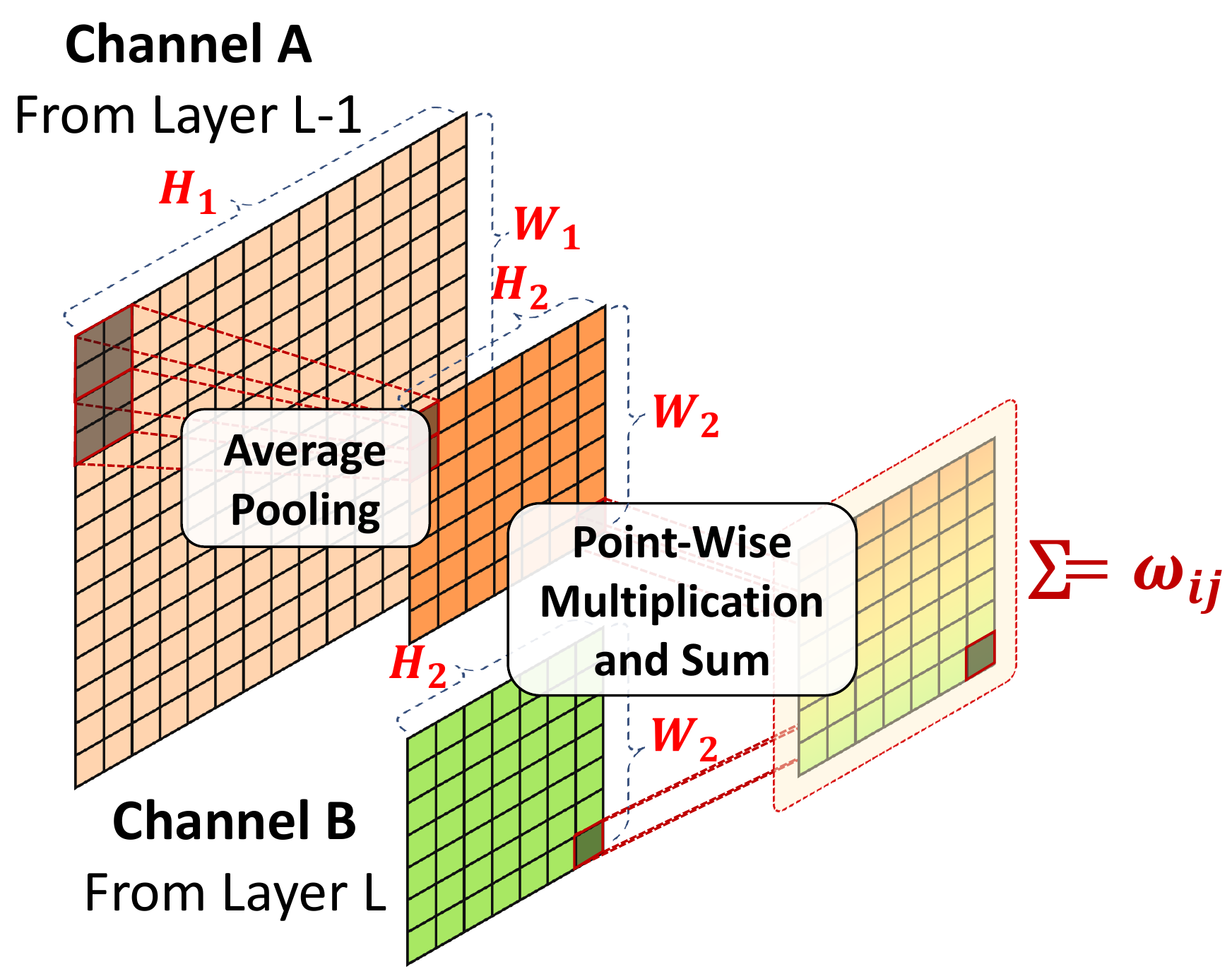}
        \caption{}
        \label{fig:edges}
    \end{subfigure}
    \caption{An illustration to show how to construct edges. (a) The basic idea of how to compute correlations. We propose to recover nodes to spatial distributions employing CAMs and regard the overlap degree of corresponding distributions as the correlation between nodes. (b) The detailed operations. We first rescale each CAMs into the same size and then compute the convolution of two normalized CAMs as the correlation value of the edge.}
    \label{fig:cam_edge}
    \vspace{-5mm}
\end{figure}

The second step is to compute edge $w_{ij}^l$ between nodes $\delta_i^l$ and $\delta_j^{l-1}$.
Since pooling operation erases the spatial information, which encodes relations between different nodes, 
we propose to utilize CAMs~\cite{zhou2016learning} of each node to recover relations as illustrated in Figure~\ref{fig:cam_edge}.
We first compute CAMs of nodes as follows:
\vspace{-3mm}
\begin{equation} \label{equ:cam}
    \mathbf{u}^l_i = \sum_{j=1}^{c^l} a_{ij} \mathbf{z}^l_j \in \mathbb{R}^{h_l \times w_l},
    \vspace{-3mm}
\end{equation}
where $\mathbf{z}^l_j$ denotes the $j$-th slice of the feature map $\mathbf{z}^l$, 
and $a_{ij}$ indicates weights of the linear layer $h^l(\cdot)$.
We consider two nodes correlated if the two distributions of the corresponding CAMs are statistically similar. 
After rescaling and vectorizing CAMs to the same scale vectors 
$\hat{\mathbf{u}}^l_i, \hat{\mathbf{u}}^{l-1}_j \in \mathbb{R}^{p}$, 
where $p=\min\{h_l,h_{l-1}\}\times \min\{w_l,w_{l-1}\}$,
we establish the correlation $\hat{\omega}_{ij}^l$ by computing the inner product of $\hat{\mathbf{u}}^l_i$ and $\hat{\mathbf{u}}^{l-1}_j$ as follows:
\vspace{-2mm}
\begin{equation}
    \hat{\omega}_{ij}^l = \langle \hat{\mathbf{u}}^l_i, \hat{\mathbf{u}}^{l-1}_j \rangle.
    \vspace{-2mm}
\end{equation}
To obtain the final edges $\omega_{ij}^l$, we adopt the momentum updating strategy to gradually incorporate all training samples:
\begin{eqnarray} \label{eq:momentum}
    \omega_{ij}^l \leftarrow \gamma \omega_{ij}^l + (1 - \gamma) \hat{\omega}_{ij}^l,
\end{eqnarray}
where $\gamma$ is a momentum factor.

\begin{figure*}[t]
    \centering
    \includegraphics[width=0.97\textwidth]{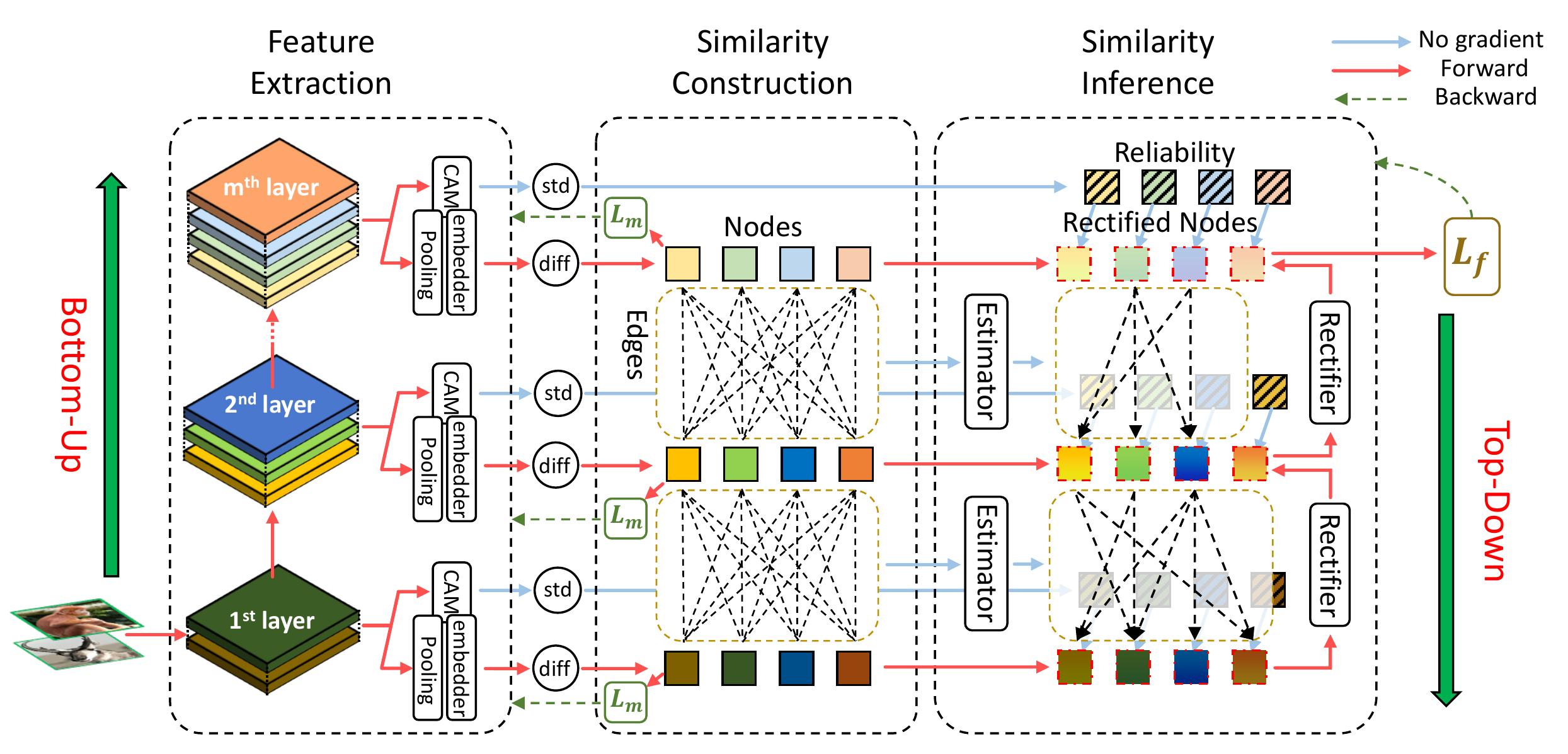}
    \caption{An illustration of the architecture of the proposed AVSL framework. We first extract a set of feature maps from multiple layers of a CNN network and perform global pooling followed by linear projection to obtain the set of embedding. We then compute the square of the absolute differences between the corresponding embeddings as similarity nodes. For similarity inference, we first estimate the reliability of the similarity nodes and rectify them using the most correlated ones in the adjacent lower level. We compute the overall similarity as the sum of the rectified similarity nodes of the top layer which can be conveniently attributed to specific similarity nodes in different levels.} 
    \label{fig:framework}
    \vspace{-4mm}
\end{figure*}

\subsection{Top-Down Similarity Inference} \label{sec:inference}

Having constructed the graph $\mathcal{H}$ composed of similarity nodes $\delta_i^l$ and correlation edges $\omega_{ij}^l$,
we want to incorporate them to compute an overall similarity.
Taking full advantage of the hierarchy consistency of CNNs, 
we propose a top-down similarity inference method based on the graph $\mathcal{H}$.
On the one hand, we argue that different levels of features encode complementary information, enabling the corresponding similarity nodes to produce relatively independent similarity judgment.
On the other hand, the most correlated similarity nodes of adjacent levels should be consistent, which can be used as a non-trivial constraint to restrict the overall similarity.
Motivated by this, we propose to identify unreliable higher-level similarity nodes and rectify them using adjacent lower-level similarity nodes with the largest correlations. 

Analogous to the process of humans comparing images from coarse to fine, we infer the overall similarity from top to bottom. 
We first estimate the reliability of the similarity nodes at the $l$-th layer to identify the unreliable ones.
Intuitively, we deem a similarity node unreliable if its corresponding CAM is unable to focus clearly on specific regions.
Formally, we employ the standard deviation of normalized CAMs to compute the reliability as: 
\begin{equation} \label{eq:reliability}
    \eta_i^l = std(\hat{\mathbf{u}}_i^l) \cdot std(\hat{\mathbf{u}}'^l_i),
\end{equation}
where $std(\cdot)$ denotes the standard deviation
and $\hat{\mathbf{u}}_i^l$, $\hat{\mathbf{u}}'^l_i$ indicate normalized CAMs of samples $x,x'$.
Then we apply a sigmoid function to map $\eta_i^l$ to the range of $(0,1)$ as:
\begin{equation} \label{equ:unreliability}
    p_i^l = \sigma(\alpha_i^l \eta^l_i + \beta_i^l)
    = \frac{e^{\alpha_i^l \eta^l_i + \beta_i^l}}{e^{\alpha_i^l \eta^l_i + \beta_i^l} + 1} \in (0,1),
\end{equation}
where $\alpha_i^l,\beta_i^l$ are node-wise learnable parameters.

We then rectify unreliable nodes at the higher-level layer with the correlated ones at the adjacent lower-level layer. 
For an unreliable $\delta^l_i$ at the $l$-th layer, we denote the index set of $k$ most correlated nodes at the $(l-1)$-th layer as:
\begin{equation} \label{topk}
    \mathbf{I}(\delta^l_i) = \{j | \omega^l_{ij} \in \mathbf{\max}_k \{\omega^l_{im:m=1,2,\cdots,r}\} \},
\end{equation}
where $\mathbf{\max}_k(\cdot)$ denotes the set of $k$ largest values.
Subsequently, we compute the rectified similarity node $\hat{\delta}^l_i$ by the sum of the original one $\delta^l_i$ and adjacent related low-level ones $\delta^{l-1}_j$ weighted by the unreliability $p^l_i$ as follows:
\begin{align} \label{equ:imp_rect}
    \hat{\delta}^l_i 
    = \left\{ 
        \begin{aligned}
            &p^l_i \delta^l_i + (1 - p^l_i) \sum_{j=1}^r \tilde{\omega}^l_{ij} \hat{\delta}^{l-1}_j,
            \!\! & 2 \leq l \leq L \\
            &\delta^l_i,
            \!\! & l = 1
        \end{aligned}
    \right.
\end{align}
where 
$\tilde{\omega}^l_{ij} = \frac{\mathbb{I}_{j \in I(\delta^l_i)} \omega^l_{ij}}{\sum_{k=1}^r \mathbb{I}_{k \in I(\delta^l_i)} \omega^l_{ik}} \in [0,1]$ denotes the normalized edges,
$\mathbb{I}_{(\cdot)}$ is the indicative function,
and $\delta^l_i, \omega^l_{ij}$ are nodes and edges of the graph $\mathcal{H}$.
Since $\delta^l_i \geq 0, p^l_i \in (0,1), \tilde{\omega}^l_{ij} \geq 0$, we know that $\hat{\delta}^l_i \geq 0$.
For convenience, we reorganize \eqref{equ:imp_rect} in a matrix format as follows:
\begin{align} \label{equ:imp_matrix}
    &\hat{\pmb{\delta}}^l
    = \mathbf{P}^l \pmb{\delta}^l + (\mathbf{I} - \mathbf{P}^l)
    \tilde{\mathbf{W}}^l \hat{\pmb{\delta}}^{l-1} \\
    &\begin{aligned}
        \text{where} ~~
        \hat{\pmb{\delta}}^l &= [\hat{\delta}^l_1 ~ \hat{\delta}^l_1 ~ \cdots ~ \hat{\delta}^l_r] \in \mathbb{R}^r \notag \\ 
        \pmb{\delta}^l &= [\delta^l_1 ~ \delta^l_2 ~ \cdots ~ \delta^l_r] \in \mathbb{R}^r \notag\\
        \mathbf{P}^l &= \left\{
        \begin{aligned}
            &\text{diag}(p^l_1,\cdots,p^l_r), \! &l \geq 2 \\
            &\mathbf{I}, \! &l=1 
        \end{aligned}
        \right. \\
        \hat{\mathbf{W}}^l &= (\tilde{\omega}^l_{ij}) \in \mathbb{R}^{r\times r}
    \end{aligned}.
\end{align}

Finally, we define the overall similarity between two images as the sum of rectified top-level similarity nodes as
$\hat{d} = \sum_{i=1}^r \hat{\delta}^L_i$,
while the smaller value indicates more similarity.
Following \eqref{equ:imp_matrix}, we can infer it recursively in a top-down manner efficiently. 

\subsection{Attributable Visual Similarity Learning} \label{sec:framework}

The proposed AVSL framework employs a bottom-up similarity construction and top-down similarity inference method based on hierarchy consistency to extend the conventional similarity learning, as shown in Figure~\ref{fig:framework}.
We divide our framework into three phases: \textbf{training}, \textbf{attribution}, and \textbf{evaluation}.

\paragraph{Training:}
In order to learn network parameters, We define $l$-th level similarity as $d^l = \sum_{i=1}^r \delta^l_i$.
The proposed AVSL is compatible with existing deep metric learning methods with various loss functions and sampling strategies to further improve their performance.
For a particular loss function $L(\cdot)$, the overall objective of the proposed AVSL framework is formulated as follows:
\vspace{-3mm}
\begin{equation}
    \min_{\theta_1, \theta_2} J = 
    \min_{\theta_1} \sum_{l=1}^L L^m(d^l) + \min_{\theta_2} L^f(\hat{d})
    \vspace{-3mm}
\end{equation}
where $\theta_1$ corresponds to the CNN network parameters 
and $\theta_2$ represents the parameters of the similarity inference module including $\alpha^l_i, \beta^l_i$ of the reliability estimatation modules in \eqref{equ:unreliability}.
We only use the loss on overall similarities to train the similarity inference module and the loss on level similarities to train the similarity construction module. 
$L^m$ targets at learning discriminative embeddings for each layer, while $L^f$ only aims at learning the similarity inference process to obtain an accurate and robust overall similarity.

\paragraph{Attribution:}
It is essential to analyze how the model infers the similarity. 
We reorganize the overall similarity $\hat{d}$ in a linear combination format following ~\eqref{equ:imp_matrix} as: 
\begin{align} \label{equ:sensitivity}
    \hat{d} &=\sum_{i=1}^r \hat{\delta}^L_i = \mathbf{1} \hat{\pmb{\delta}}^L \\
    &= \mathbf{1} \mathbf{P}^L \pmb{\delta}^L + \mathbf{1} (\mathbf{I} - \mathbf{P}^L)
    \tilde{\mathbf{W}}^L \hat{\pmb{\delta}}^{L-1} \notag \\
    &= \sum_{l=1}^L \mathbf{1} \mathbf{\Lambda}^l \pmb{\delta}^l
    = \sum_{l=1}^{L} \sum_{i=1}^r \lambda^l_i \delta^l_i, \notag
\end{align}
where 
$\mathbf{\Lambda}^l = (\mathbf{I} - \mathbf{P}^L) \tilde{W}^L \cdots (\mathbf{I} - \mathbf{P}^{l+1}) \tilde{W}^{l+1} \mathbf{P}^l$,
and $\pmb{\lambda}^l = [\lambda^l_1 ~ \lambda^l_1 ~ \cdots \lambda^l_r] = \mathbf{1}^T \mathbf{\Lambda}^l$.
The weight $\lambda^l_i$ represents the sensitivity of the overall similarity to each similarity node $\delta^l_i$,
which means the change of the similarity node $\delta^l_i$ will contribute more to the change of the overall similarity $\hat{d}$
if its corresponding weight $\lambda^l_i$ is larger.
Finally, saliency maps are generated to demonstrate the attribution process.
The proposed AVSL framework is convenient for visualization
since we compute similarity nodes and corresponding CAMs simultaneously in a single forward propagation.

\paragraph{Evaluation:}
During the evaluation, we freeze all parameters and only compute the overall similarities $\hat{d}$ using \eqref{equ:sensitivity} to represent the similarities between the given query samples and gallery samples.

\section{Experiment} \label{sec:exp}

\begin{table*}[htbp] \small
\centering
\caption{Recall@K(\%) on the test sets of CUB-200-2011, Cars196, and Stanford Online Products.}
\begin{tabular}{lccccc|cccc|ccc}
\toprule
Datasets &   & \multicolumn{4}{c}{CUB-200-2011} & \multicolumn{4}{c}{Cars196} & \multicolumn{3}{c}{Stanford Online Products} \\
\cmidrule(lr){3-6} 
\cmidrule(lr){7-10} 
\cmidrule(lr){11-13}
Methods & Setting & R@1 & R@2 & R@4 & R@8 & R@1 & R@2 & R@4 & R@8 & R@1 & R@10 & R@100 \\
\midrule
HDC~\cite{yuan2017hard} & 384G 
& 53.6 & 65.7 & 77.0 & 85.6  
& 73.7 & 83.2 & 89.5 & 93.8
& 70.1 & 84.9 & 93.2 \\
DAML~\cite{duan2018deep} & 512BN 
& 52.7 & 65.4 & 75.5 & 84.3  
& 75.1 & 83.8 & 89.7 & 93.5
& 68.4 & 83.5 & 92.3 \\
DVML~\cite{lin2018deep} & 512BN 
& 52.7 & 65.1 & 75.5 & 84.3  
& 82.0 & 88.4 & 93.3 & 96.3
& 70.2 & 85.2 & 93.8 \\
Angular~\cite{wang2017deep} & 512G 
& 53.6 & 65.0 & 75.3 & 83.7  
& 71.3 & 80.7 & 87.0 & 91.8
& 67.9 & 83.2 & 92.2 \\
DAMLRRM~\cite{xu2019deep} & 512G 
& 55.1 & 66.5 & 76.8 & 85.3  
& 73.5 & 82.6 & 89.1 & 93.5
& 69.7 & 85.2 & 93.2 \\
DE-DSP~\cite{duan2019deep} & 512G 
& 53.6 & 65.5 & 76.9 & - 
& 72.9 & 81.6 & 88.8 & -
& 68.9 & 84.0 & 92.6 \\
HDML~\cite{zheng2019hardness} & 512BN 
& 53.7 & 65.7 & 76.7 & 85.7  
& 79.1 & 87.1 & 92.1 & 95.5
& 68.7 & 83.2 & 92.4 \\
A-BIER~\cite{opitz2018deep} & 512G 
& 57.5 & 68.7 & 78.3 & 86.2  
& 82.0 & 89.0 & 93.2 & 96.1
& 74.2 & 86.9 & 94.0 \\
ABE~\cite{kim2018attention} & 512G 
& 60.6 & 71.5 & 79.8 & 87.4 
& 85.2 & 90.5 & 94.0 & 96.1
& 76.3 & 88.4 & 94.8 \\
MS~\cite{wang2019multi} & 512BN 
& 65.7 & 77.0 & 86.3 & 91.2 
& 84.1 & 90.4 & 94.0 & 96.5
& 78.2 & 90.5 & 96.0 \\
SoftTriple~\cite{qian2019softtriple} & 512BN 
& 65.4 & 76.4 & 84.5 & 91.6 
& 86.1 & 91.7 & 95.0 & 97.3
& 78.3 & 90.3 & 95.9 \\
Circle~\cite{sun2020circle} & 512BN 
& 66.7 & 77.4 & 86.2 & 91.2 
& 83.4 & 89.8 & 94.1 & 96.5
& 78.3 & 90.5 & 96.1 \\
DCML~\cite{zheng2021deep1} & 512R
& 68.4 & 77.9 & 86.1 & 91.7
& 85.2 & 91.8 & 96.0 & 98.0
& 79.8 & 90.8 & 95.8 \\
DIML~\cite{zhao2021towards} & 512R
& 68.2 & - & - & - 
& 87.0 & - & - & -
& 79.3 & - & - \\
DRML~\cite{zheng2021deep2} & 512R
& 68.7 & 78.6 & 86.3 & 91.6
& 86.9 & 92.1 & 95.2 & 97.4
& \textcolor[rgb]{1,0,0}{79.9} & 90.7 & 96.1 \\
\midrule
Margin~\cite{wu2017sampling} & 512R 
& 65.6 & 75.9 & 84.3 & 90.8 
& 78.2 & 86.7 & 92.3 & 95.3
& 72.4 & 85.3 & 92.8 \\ 
Margin-AVSL & 512R 
& \textbf{68.8} & \textbf{79.2} & \textbf{87.3} & \textbf{92.7} 
& \textbf{81.1} & \textbf{88.8} & \textbf{93.4} & \textbf{96.4}
& \textbf{76.8} & \textbf{89.2} & \textbf{95.4} \\
\midrule
ProxyAnchor~\cite{kim2020proxy} & 512R 
& 69.7 & 80.0 & 87.0 & 92.4 
& 87.7 & 92.9 & 95.8 & 97.9
& 78.4 & 90.5 & 96.2 \\
ProxyAnchor-AVSL & 512R 
& \textcolor[rgb]{1,0,0}{\textbf{71.9}} & \textcolor[rgb]{1,0,0}{\textbf{81.7}} & \textcolor[rgb]{1,0,0}{\textbf{88.1}} & \textcolor[rgb]{1,0,0}{\textbf{93.2}} 
& \textcolor[rgb]{1,0,0}{\textbf{91.5}} & \textcolor[rgb]{1,0,0}{\textbf{95.0}} & \textcolor[rgb]{1,0,0}{\textbf{97.0}} & \textcolor[rgb]{1,0,0}{\textbf{98.4}}
& \textbf{79.6} & \textcolor[rgb]{1,0,0}{\textbf{91.4}} & \textcolor[rgb]{1,0,0}{\textbf{96.4}} \\
\bottomrule
\end{tabular}
\label{tab:Recall}
\vspace{-3mm}
\end{table*}
    
In this section, we conducted experiments on three widely used datasets including 
CUB-200-2011~\cite{wah2011caltech}, 
Cars196~\cite{krause20133d}, 
and Stanford Online Products~\cite{oh2016deep}
to evaluate the accuracy and interpretability of our AVSL framework. 
We used the Recall@Ks as the performance metrics, which compute the percentage of well-separated samples acknowledged if we can find at least one corrected retrieved sample in the K nearest neighbors.

\subsection{Dataset} \label{sec:dataset}

For quantitative evaluation, we conducted experiments under a zero-shot setting following the non-intersecting dataset partition protocol~\cite{oh2016deep}. 
The split scheme of datasets are as follows:
\begin{itemize}
    \item \textbf{CUB-200-2011}~\cite{wah2011caltech} consists of 200 bird species and 11,788 images. 
    We split the first 100 species (5,864 images) for training and the rest 100 species (5,924 images) for testing.
    \vspace{-2mm}
    \item \textbf{Cars196}~\cite{krause20133d} contains 196 car types and 16,185 images. The first 98 types (8,054 images) were used for training while the other 98 types (8,131 images) were kept for testing. 
    \vspace{-2mm}
    \item \textbf{Stanford Online Products}~\cite{oh2016deep} includes 22,634 classes of online products totaling 120,053 images. We divide the first 11,318 classes (59,551 images) into training set and the rest 11,316 classes (60,502 images) into testing set.
\end{itemize}

For qualitative demonstration, we further visualized the similarity attribution results of some randomly selected samples in CUB-200-2011 and Cars196. 
All datasets are publicly available for non-commercial research and educational purposes.

\subsection{Implementation Details} \label{sec:implement}
We conducted all the experiments using the PyTorch package~\cite{paszke2019pytorch} on an NVIDIA RTX 3090 GPU and employed the ResNet50~\cite{he2016deep} as the CNN feature extractor (i.e., $f^m$) for fair comparisons. 
Limited by the GPU device memory, we only selected feature maps for every three layers for similarity construction (i.e., layers 3, 4, and 5). 
We employed a global pooling operation (i.e., $g^l$) and a linear layer (i.e., $h^l$) after each selected layer. 
We fixed the embedding size to 512 for all selected layers.
For data argumentation, we first resized images to 256 by 256 to apply random reshaping and horizontal flip and then randomly cropped them to 224 by 224.
Before training, we initialized the CNN with weights pre-trained on ImageNet ILSVRC dataset~\cite{russakovsky2015imagenet}.
We adopted AdamW~\cite{loshchilov2017decoupled} to train our model with an initial learning rate $1\times 10^{-4}$ and a weight decay of $0.0001$. 
We fixed the batch size to 180 and set the momentum factor $\gamma$ to 0.5.
For the margin loss~\cite{wu2017sampling}, we set the margin factors $\alpha$ and $\beta$ to 1.2 and 0.2, respectively.
For the ProxyAnchor loss~\cite{kim2020proxy}, we set the temperature $\alpha=16$, positive margin $\gamma_{pos}=1.8$, and negative margin $\gamma_{neg}=2.2$.
We tuned all hyperparameters by grid search on a reserved validation set.

\subsection{Quantitative Results and Analysis}

\paragraph{Comparisons with existing methods:}
We applied the proposed AVSL framework to the margin loss~\cite{wu2017sampling} and the ProxyAnchor loss~\cite{kim2020proxy} for demonstration and compared our framework with several baseline methods.
Table \ref{tab:Recall} shows the image retrieval performance on the CUB-200-2011~\cite{wah2011caltech}, Cars196~\cite{krause20133d}, and Stanford Online Products~\cite{oh2016deep} respectively. 
We mark the best results with bold red and highlight our superior results over the associated methods without AVSL in bold black.

We observe that our AVSL framework can greatly improve the original deep metric learning methods by a large margin and achieve state-of-the-art performance on three datasets.
We ascribe the improvement to exploiting the graph structure by employing the hierarchy consistency between different similarity nodes as the inductive bias which is consistent with how humans perceive the semantic visual similarity. 
By rectifying unreliable higher-level similarity nodes with the most correlated ones in the lower-level layer, we achieve a more accurate and robust similarity measure with the proposed top-down similarity inference.

\begin{table}[tb] \small
\centering
\caption{Ablation study with different model settings.}
\begin{tabular}{lcccc}
\toprule
Method & R@1 & R@2 & R@4 & R@8 \\
\midrule
ProxyAnchor & 87.7 & 92.9 & 95.8 & 97.9 \\
ProxyAnchor + M & 89.7  & 93.9  & 96.3  & 97.9  \\
ProxyAnchor + M \& R & 89.9  & 94.0  & 96.4  & 98.1  \\
ProxyAnchor + M (concat) & 90.6  & 94.6  & 96.8  & 98.2  \\
\midrule
\textbf{ProxyAnchor + AVSL} & \textbf{91.5} & \textbf{94.8} & \textbf{96.9} & \textbf{98.4} \\
\bottomrule
\end{tabular}
\vspace{-4mm}
\label{tab:ablation}
\end{table}

\vspace{-4mm}
\paragraph{Ablation study:}
We first performed an ablation study to evaluate the contribution of each component of the proposed AVSL framework.
We report the experimental results on the Cars196~\cite{krause20133d} dataset with the ProxyAnchor loss~\cite{kim2020proxy}, as shown in Table \ref{tab:ablation}, 
but we observe similar outcomes with the other loss functions. 
We highlight the best results using bold numbers.

\textbf{ProxyAnchor} denotes the baseline method of using the ProxyAnchor loss. 
\textbf{+ M} is short for `multi-layer'. 
In this trial, we exerted extra loss constraints on embeddings of all layers (i.e., layers 3, 4, and 5).
\textbf{+ R} stands for `reliability', 
which means that we utilize the reliabilities defined in ~\eqref{equ:unreliability} and only keep reliable nodes to compute the similarities.
Based on \textbf{+ M \& R} setting, if we further consider edges between similarity nodes, we can get complete components of the proposed AVSL framework.
In the \textbf{+ M (concat)} setting, we concatenated embeddings of all three layers (i.e., the final dimension equals to $3 \times 512 = 1536$) to compute similarities,
which is a strong baseline to further demonstrate the effectiveness of the proposed AVSL framework.

We observe that the proposed AVSL framework achieves better performance than all the compared counterparts and all modules contribute to the overall improvement.
In particular, imposing loss constraints on the embeddings of hidden layers can boost the performance of the original method by 2.0\%.
It is also beneficial to employ reliabilities defined in ~\eqref{equ:unreliability} to guide the selection of informative nodes.
Subsequently, exploiting the relations among similarity nodes and employing the hierarchy consistency for similarity inference can further improve the performance by 1.6\%. 
We also see that our method suppresses \textbf{+ M (concat)}, which demonstrates that learning informative relations and reliabilities is crucial for effective inference.

\begin{figure}[tb]
    \centering
    \begin{subfigure}{0.495\linewidth}
        \includegraphics[width=1\linewidth]{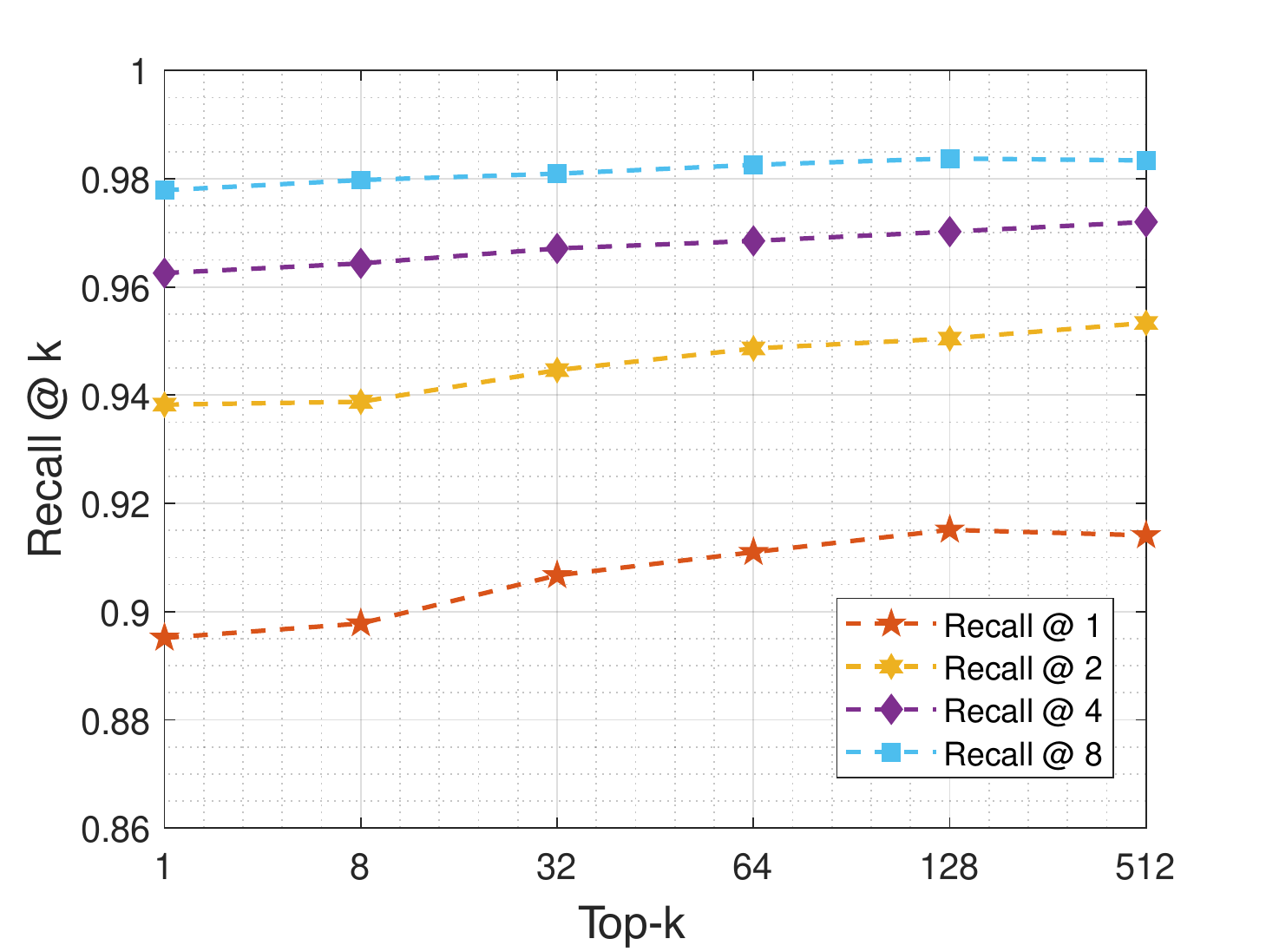}
        \caption{Different top-k values.}
        \label{fig:topk_value}
    \end{subfigure}
    \begin{subfigure}{0.495\linewidth}
        \includegraphics[width=1\linewidth]{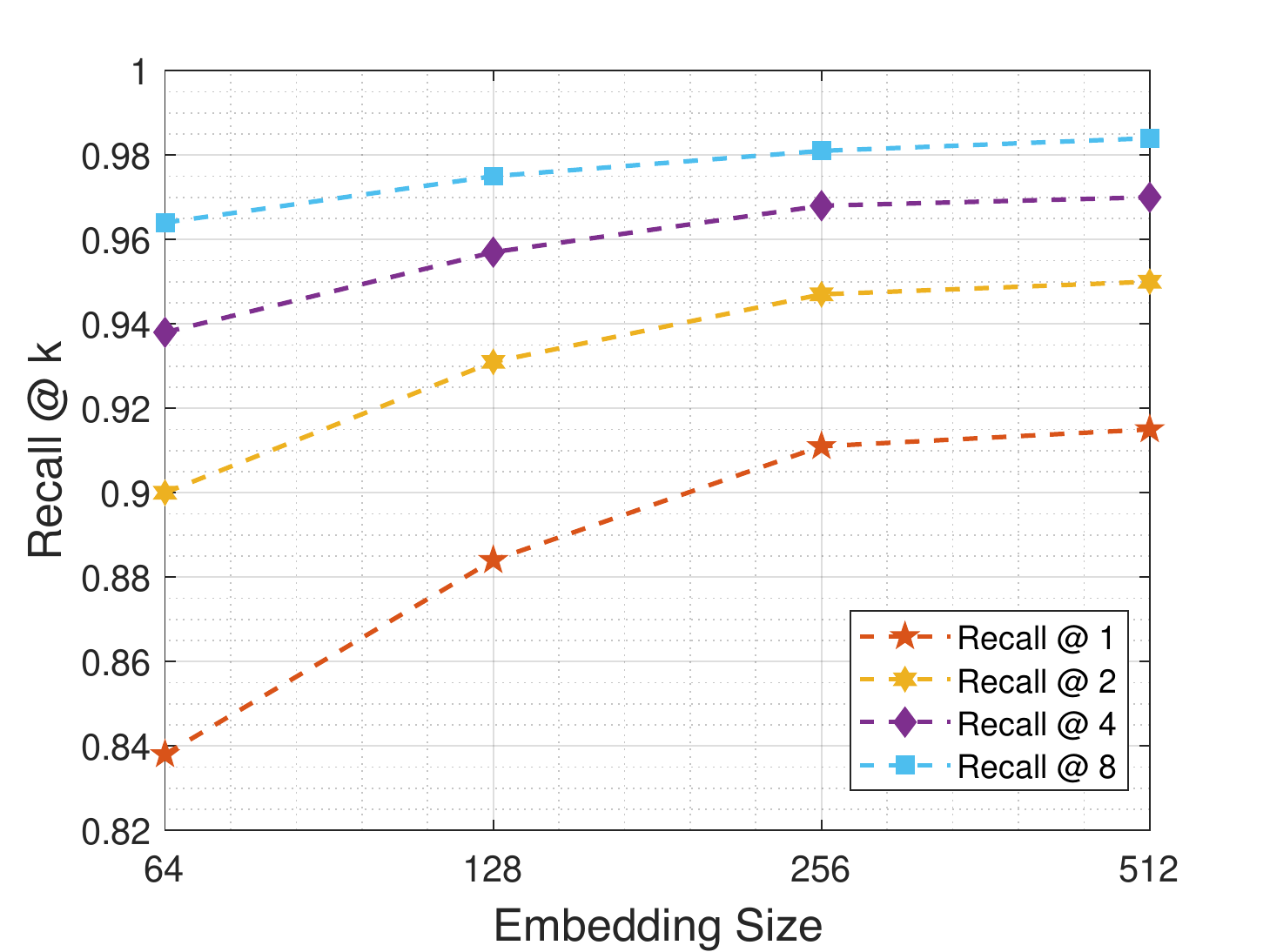}
        \caption{Different embedding sizes.}
        \label{fig:embed_size}
    \end{subfigure}
    \caption{Influence of hyperparameters.}
    \vspace{-5mm}
\end{figure}

\vspace{-4mm}
\paragraph{Influence of hyperparameters:}

\begin{figure}[t]
\centering
\includegraphics[width=0.95\linewidth]{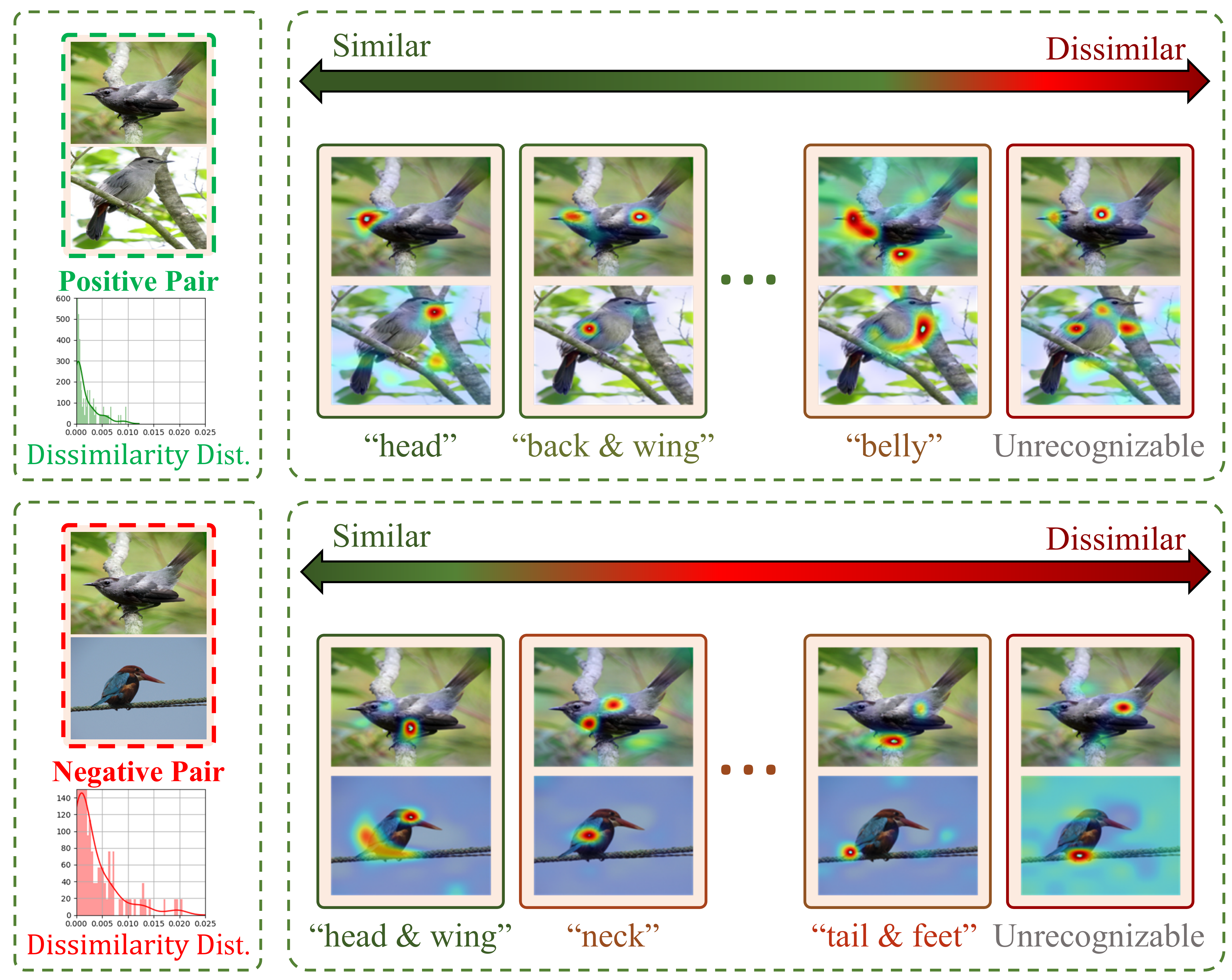}
\caption{Visualization of the attribution result.
We randomly select a triplet from CUB-200-2011 and rank the results according to the similarity among the 128 most reliable nodes for each sample pair. We use green and red boxes to denote positive and negative pairs, respectively.
Best viewed in color.
} 
\label{fig:visualization_attr}
\vspace{-4mm}
\end{figure}

The k value defined in ~\eqref{topk} controls how many adjacent related nodes participate in rectification.
Figure~\ref{fig:topk_value} reveals the continuous improvement when increasing the k value.
And we can further discover that the influence of k shows diminishing marginal effects, 
thus we fixed k to 128 to complete all other experiments.
In addition, the dimension of embeddings significantly impacts the performance as shown in Figure~\ref{fig:embed_size}, 
and larger embedding size leads to higher performance.
In particular, when only fixing the dimension to 128, our proposed AVSL could surpass all other methods with a recall@1 score of 88.4\% on the Cars196 dataset,
which further demonstrates the effectiveness of our framework.

\vspace{-2mm}
\subsection{Visualization}

To verify the interpretability of the proposed AVSL framework,
we randomly selected a triplet from CUB-200-2011 to show the attribution results, as shown in Figure~\ref{fig:visualization_attr}.
For the pairs in the triplet, we first selected the 128 most reliable similarity nodes,
and then ranked those nodes according to their similarities.
We observe that most of the CAMs of nodes focus on specific parts of images while some saliency maps are unrecognizable.
We think that this phenomenon is due to the singularity of the relationships between spatial coordinates and concepts.
Also, we discover that the dissimilarity distribution of the negative pair is more dispersed than the positive one as shown on the left of Figure~\ref{fig:visualization_attr}.
This means that the nodes of the negative pair are more likely to be dissimilar, 
which is beneficial to classifying samples from different classes.

\begin{figure}[t]
\centering
\includegraphics[width=0.95\linewidth]{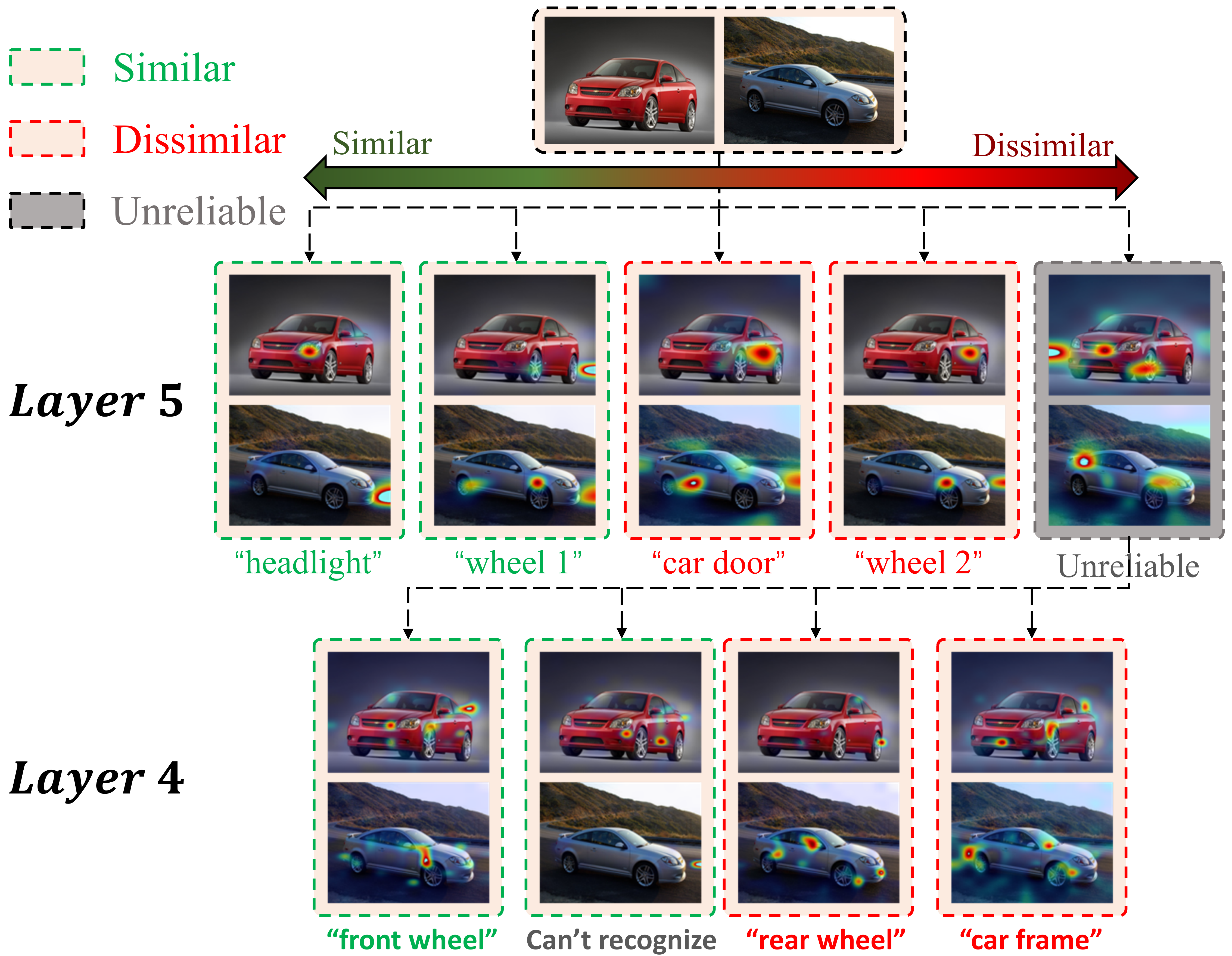}
\caption{Visualization of the attribution process.
We randomly select 2 samples from Cars196 and attribute the overall similarity to the specific similarity nodes in an undirected graph manner. 
Best viewed in color.
} 
\label{fig:visualization_infer}
\vspace{-4mm}
\end{figure}

To further understand the underlying mechanism of the inference process, 
we also randomly selected a sample pair from Cars196 for similarity attribution, as shown in Figure~\ref{fig:visualization_infer}
\footnote{More detailed inference graphs of samples from both CUB-200-2011 and Cars196 are included in the appendix.}.
From top to bottom, we first selected top-128 reliable nodes with high $p^l_i$ scores among 512 nodes and further displayed the two most similar nodes framed in the green dotted box as well as the two most dissimilar ones framed in the red dotted box. 
Subsequently, we decompose one unreliable node into adjacent related nodes.
We observe that similarity nodes with higher sensitivity value $\lambda^l_i$ are more likely positioned in a higher layer and correspond to clearer concepts such as ``headlight", ``wheel", and ``door" and low-level saliency maps are difficult to distinguish concepts.
This demonstrates that high features tend to encode object-level patterns while low features focus on pixel-level patterns.
In addition, we discover that nodes and concepts may not correspond to each other one-to-one.
For example, multiple nodes may focus on the ``wheel" part of cars, 
which indicates that concepts extracted by CNNs are not well disentangled.

\vspace{-2mm}
\section{Limitations}

During evaluation, our AVSL framework needs to maintain a similarity matrix with the spatial complexity of $O(N^2)$ to compute the similarity between two images, where $N$ denotes the number of samples. 
When dealing with large-scale datasets, we use computation tricks such as matrix slicing to reduce the memory usage to achieve partially parallel computing.
This also affects training of proxy-based methods (e.g., the ProxyAnchor loss) when the number of classes are large.
On the Stanford Online Products dataset, it is impossible to maintain the similarities between samples in a mini-batch and 11,318 proxies simultaneously on a 24GB-memory device.
We thus tailor the loss to only constrain the similarities between samples and positive proxies, which may lead to inferior performance. 

\vspace{-2mm}
\section{Conclusion}

In this paper, we have presented an attributable visual similarity learning (AVSL) framework to learn a more accurate and interpretable similarity. 
We adopt a hierarchy consistency as the inductive bias and employ a bottom-up similarity construction and top-down similarity inference method to model the visual similarity, which first estimates the reliability of similarity nodes at a higher level and then rectifies the unreliable ones using the correlated ones in the adjacent lower level.
We have conducted experiments on three widely used datasets to demonstrate the superiority of our framework on both accuracy and interpretability.
While our framework is motivated by human visual similarity perception, we believe it can also be adapted to other modalities of information such as text and speech for better interpretability, which is an interesting future work.

\vspace{-1mm}
\section*{Acknowledgement}
This work was supported in part by the National Natural Science Foundation of China under Grant 62125603 and Grant U1813218, and in part by a grant from the Beijing Academy of Artificial Intelligence (BAAI).

\clearpage
\appendix
\section*{\LARGE Appendix}
\section{Implementation Details}

\begin{figure*}[t]
\centering
\includegraphics[width=1\linewidth]{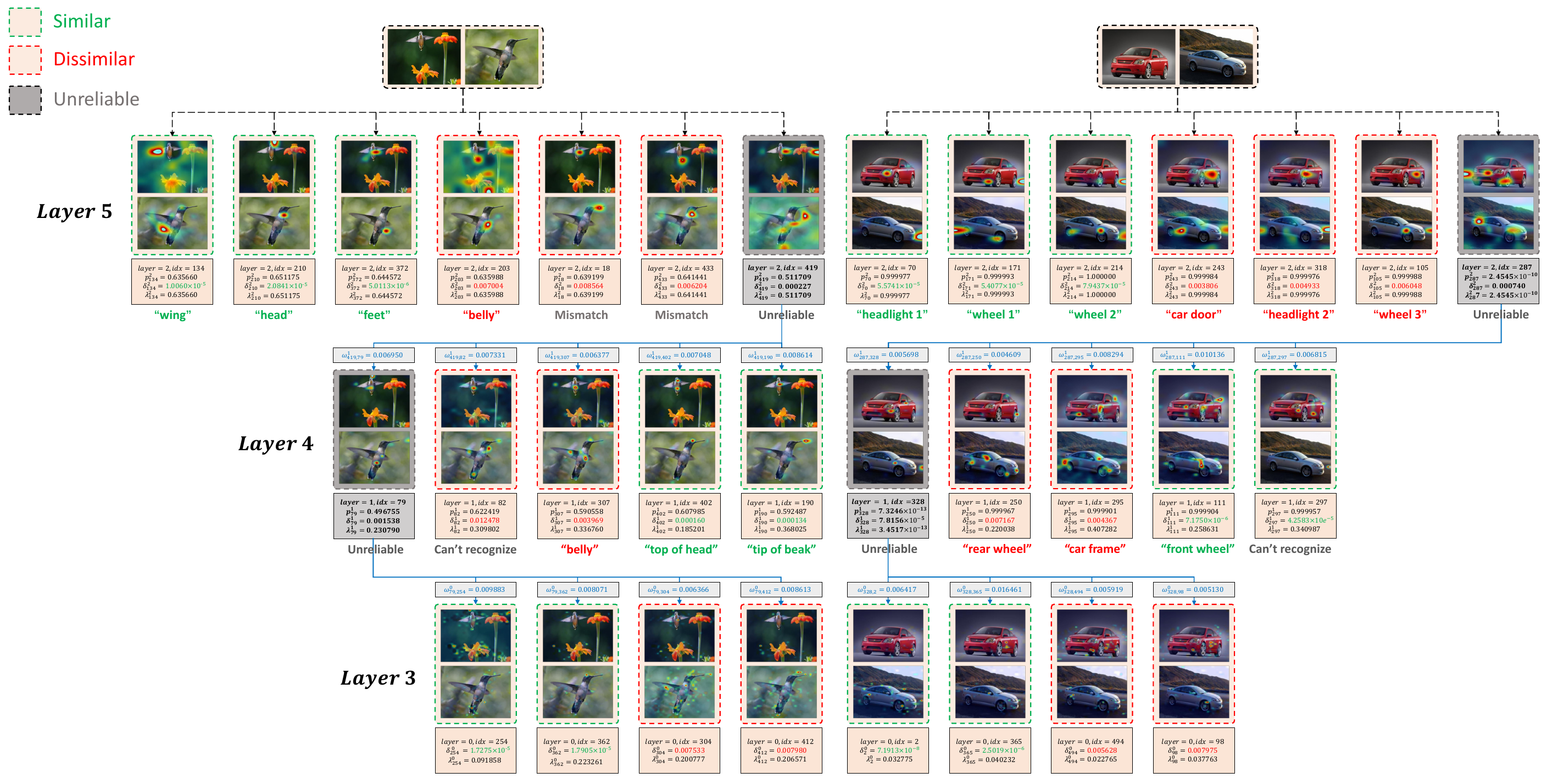}
\caption{Visualization of the similarity inference and attribution.
We randomly selected two sample pairs from CUB-200-2011 and Cars196.
For each pair of images, we attribute the overall similarity to the specific similarity nodes in an undirected graph. We report the corresponding values of reliabilities, nodes, and sensitivities under each node. 
Best viewed in color.
} 
\label{fig:visualization}
\vspace{-4mm}
\end{figure*}

\subsection{Loss Functions}

Loss functions in deep metric learning can be categorized into pair-based methods and proxy-based methods. 
During training, we apply the proposed AVSL framework to the margin loss~\cite{wu2017sampling} and the ProxyAnchor loss~\cite{kim2020proxy} as the representative pair-based and proxy-based methods, respectively, to verify the effectiveness.

\paragraph{Margin loss~\cite{wu2017sampling}}
 compresses positive pairs while repelling negative pairs in the embedding space as follows:
\begin{align} \label{equ:margin}
    L_{margin} = &\frac{1}{\vert \mathbf{P} \vert} 
    \sum_{(x,x^+)\in \mathbf{P}} [d(x,x^+) - (\beta_{y_x} - \alpha)]_+ \notag \\
    &\frac{1}{\vert \mathbf{N} \vert} 
    \sum_{(x,x^-)\in \mathbf{N}} [(\beta_{y_x} + \alpha) - d(x,x^-)]_+,
\end{align}
where $[\cdot]_+$ is the hinge function (i.e., $[x]_+ = \max\{x,0\}$) and $d(\cdot,\cdot)$ denotes the Euclidean distance. 
We use $\mathbf{P}$ and $\mathbf{N}$ to indicate the set of positive pairs and negative pairs and $\vert \cdot \vert$ to denote the set of the size.
To address the variable intra-class distributions, the margin loss introduce a learnable parameter $\pmb{\beta} \in \mathbb{R}^C$ to adaptively control the range of each class, where $C$ denotes the number of classes.
$\alpha$ is a fixed parameter to enforce a large margin between classes.

\paragraph{ProxyAnchor loss~\cite{kim2020proxy}}
instead constrains the relations between proxies and samples as follows:
\begin{align} \label{equ:proxyanchor}
    L_{pa} = &\frac{1}{\vert \mathbf{P}^+ \vert} \sum_{p\in \mathbf{P}^+} 
    \log\left(1 + \sum_{x\in \mathbf{X}_p^+} e^{-\alpha(s(x,p) - \delta)}\right) + \notag \\
    &\frac{1}{\vert \mathbf{P} \vert} \sum_{p\in \mathbf{P}}
    \log\left(1 + \sum_{x\in \mathbf{X}_p^-} e^{\alpha(s(x,p) + \delta)}\right) ,
\end{align}
where $\alpha$ is a scaling factor, $\delta$ is the margin, $s(\cdot,\cdot)$ is the cosine similarity function, $\mathbf{P}$ denotes the proxy set, and $\mathbf{P}^+$ denotes the positive proxy set where each proxy has at least one positive samples in the current batch.
Also, $\mathbf{X}_p^+$ includes the positive samples for a proxy $p$ and $\mathbf{X}_p^-$ contains the rest negative samples in the batch.
However, the original form of ProxyAnchor loss \eqref{equ:proxyanchor} is defined with the cosine similarity, while our proposed AVSL is defined in the context of dissimilarity.
To address this, we reformulate the ProxyAnchor loss as follows:
\begin{align} \label{equ:proxyanchor2}
    L_{pa} = &\frac{1}{\vert \mathbf{P}^+ \vert} \sum_{p\in \mathbf{P}^+} 
    \log\left(1 + \sum_{x\in \mathbf{X}_p^+} e^{\alpha(d(x,p) - (\beta-\tau))}\right)  + \notag \\
    &\frac{1}{\vert \mathbf{P} \vert} \sum_{p\in \mathbf{P}}
    \log\left(1 + \sum_{x\in \mathbf{X}_p^-} e^{-\alpha(d(x,p) - (\beta+\tau))}\right),
\end{align}
where $d(\cdot,\cdot)$ indicates the dissimilarity and $\beta$ and $\tau$ control the interclass margin similar to $\delta$ in \eqref{equ:proxyanchor}.

\subsection{Pooling Linearization}

To construct the similarity graph, we first employ a CNN to extract the feature map
$\mathbf{z} = f(x)$ at each level
and then reduce the feature map to a feature vector using pooling operation as
$\mathbf{v} = g(\mathbf{z})$.
Specifically, we use both max pooling $g_{max}$ and average pooling $g_{avg}$ operations following~\cite{kim2020proxy} as follows:
\begin{equation}
    v_i = \max_{h,w} z_{ihw}
    + \frac{1}{HW}\sum_{h=1}^H \sum_{w=1}^W z_{ihw}.
\end{equation}
Finally, we adopt a linear layer $h$ to map $\mathbf{v}$ into an embedding space as:
\begin{equation} \label{equ:extract}
    \mathbf{e} = h(\mathbf{v}) = (h\circ g)(\mathbf{z}).
\end{equation}
In addition, we need to compute CAMs~\cite{zhou2016learning} as follows:
\begin{equation}
    u_{ihw} = h(\mathbf{z}_{\cdot hw}) = \sum_{j=1}^c a_{ij} z_{jhw},
\end{equation}
where $c$ is the number of channels and $a_{ij}$ indicates the weights of the linear layer $h$. 
In order to maintain the spatial information of $\mathbf{e}$,
we want to ensure the following property:
\begin{equation} \label{equ:cam}
    g(\mathbf{u}) = (g\circ h)(\mathbf{z}) = \mathbf{v}.
\end{equation}
However, the pooling operation and the linear mapping is not commutative (i.e., $g\circ h \neq h \circ g$) 
since the pooling operation is a nonlinear function.
To address this, we propose a linearization operation $\tilde{g}$ as follows:
\begin{equation}
    \tilde{\mathbf{z}}_i = \tilde{g}(\mathbf{z}_i) = 
    \left\{
    \begin{aligned}
        &K\cdot z_{ihw}, &~~\text{if} ~ z_{ihw} = \max_{kl}z_{ikl} \\
        &0, &~~\text{Otherwise}
    \end{aligned}
    \right.
\end{equation}
where $K = \frac{HW}{\#\{z_{ihw} | z_{ihw} = \max_{kl} z_{ikl}\}}$.
Thus, we can decompose the pooling operation as follows:
\begin{align}
    g &= g_{max} + g_{avg} = g_{avg}\circ \tilde{g} + g_{avg} \notag \\
    &= g_{avg} \circ (\tilde{g} + \mathbb{I}),
\end{align}
where $\mathbb{I}$ denotes the identity mapping.
By employing this linearization trick, we first preprocess the feature map as follows:
\begin{equation}
    \tilde{\mathbf{z}} = (\tilde{g} + \mathbb{I})(\mathbf{z}) = \tilde{g}(\mathbf{z}) + \mathbf{z}.
\end{equation}
We then rewrite \eqref{equ:extract} and \eqref{equ:cam} as:
\begin{align}
    \mathbf{e} &= (h\circ g)(\mathbf{z}) = (h\circ g_{avg})(\tilde{\mathbf{z}}) \notag \\
    g_{avg}(\mathbf{u}) &= g_{avg}(h(\tilde{\mathbf{z}})) = (g_{avg}\circ h)(\tilde{\mathbf{z}}), \notag
\end{align}
The $g_{avg}$ and $h$ are now commutative (i.e., $g_{avg}\circ h = h\circ g_{avg}$) so that the CAMs can preserve the spatial information of the embeddings.

\section{Attribution Property}

The proposed AVSL can attribute the overall similarity to specific similarity nodes quantitively as:
\begin{align} \label{equ:sensitivity}
    \vspace{-2mm}
    \hat{d} &=\sum_{i=1}^r \hat{\delta}^L_i = \mathbf{1} \hat{\pmb{\delta}}^L
    = \mathbf{1} \mathbf{P}^L \pmb{\delta}^L + \mathbf{1} (\mathbf{I} - \mathbf{P}^L)
    \tilde{\mathbf{W}}^L \hat{\pmb{\delta}}^{L-1} \notag \\
    &= \sum_{l=1}^{L} \sum_{i=1}^r \lambda^l_i \delta^l_i,
\end{align}
where $\hat{d}$ is the overall similarity, $\delta^l_i$ is the similarity node, and $\lambda^l_i$ denotes the sensitivity of the corresponding node. 
$\lambda^l_i$ represents the influence of the corresponding node on the overall similarity.
The sensitivities have the following property:
\begin{property} \label{property}
    The sum of $\lambda^l_i$ of all nodes is a constant.
\end{property}

\begin{proof}
    We rewrite ~\eqref{equ:sensitivity} as follows:
    \begin{align}
        \hat{d} = \sum_{l=1}^L \mathbf{1} \mathbf{\Lambda}^l \pmb{\delta}^l,
    \end{align}
    where $\mathbf{\Lambda}^l = (\mathbf{I} - \mathbf{P}^L) \tilde{W}^L \cdots (\mathbf{I} - \mathbf{P}^{l+1}) \tilde{W}^{l+1} \mathbf{P}^l$,
    and $\pmb{\lambda}^l = [\lambda^l_1 ~ \lambda^l_1 ~ \cdots \lambda^l_r] = \mathbf{1}^T \mathbf{\Lambda}^l$.
    Let $\tilde{\mathbf{\Lambda}}^l = (\mathbf{I} - \mathbf{P}^L) \tilde{W}^L \cdots (\mathbf{I} - \mathbf{P}^{l+1}) \tilde{W}^{l+1}$.
    Since $\tilde{W}^l$ is normalized by row (i.e., $\tilde{W}^l \mathbf{1} = \mathbf{1}$), we can derive that:
    \begin{align}
        &(\mathbf{\Lambda}^{l+1} + \tilde{\mathbf{\Lambda}}^l) \mathbf{1} \notag \\
        = &(\mathbf{I} - \mathbf{P}^L) \tilde{W}^L \cdots (\mathbf{I} - \mathbf{P}^{l+2}) \tilde{W}^{l+2} \notag \\
        &\left(
            \mathbf{P}^{l+1}\mathbf{1} + 
            (\mathbf{I} - \mathbf{P}^{l+1})\tilde{W}^{l+1} \mathbf{1}
        \right) \notag \\
        = &(\mathbf{I} - \mathbf{P}^L) \tilde{W}^L \cdots (\mathbf{I} - \mathbf{P}^{l+2}) \tilde{W}^{l+2} \mathbf{1} \notag \\
        = &\tilde{\mathbf{\Lambda}}^{l+1} \mathbf{1}
    \end{align}
    Therefore, the sum of $\lambda^l_i$ is computed as:
    \begin{align}
        &\sum_{l=1}^L \sum_{i=1}^r \lambda^l_i 
        = \sum_{l=1}^L \mathbf{1}^T \mathbf{\Lambda}^l \mathbf{1} \notag \\
        = &\mathbf{1}^T \left(
            \sum_{l=2}^L \mathbf{\Lambda}^l \mathbf{1} + \tilde{\mathbf{\Lambda}}^1 \mathbf{1}
        \right) \notag \\
        = &\mathbf{1}^T \tilde{\mathbf{\Lambda}}^L \mathbf{1}
        = \mathbf{1}^T \mathbf{1} 
        = r,
    \end{align}
    where $r$ is the dimension of embeddings.
\end{proof}

Property~\ref{property} ensures that the absolute value of the sensitive $\lambda^l_i$ is meaningful across samples and can directly indicate the significance of the corresponding similarity node when inferring the overall similarity.

\section{More Experimental Results}

\begin{figure}[t]
    \centering
    \vspace{-3mm}
    \includegraphics[width=1\linewidth]{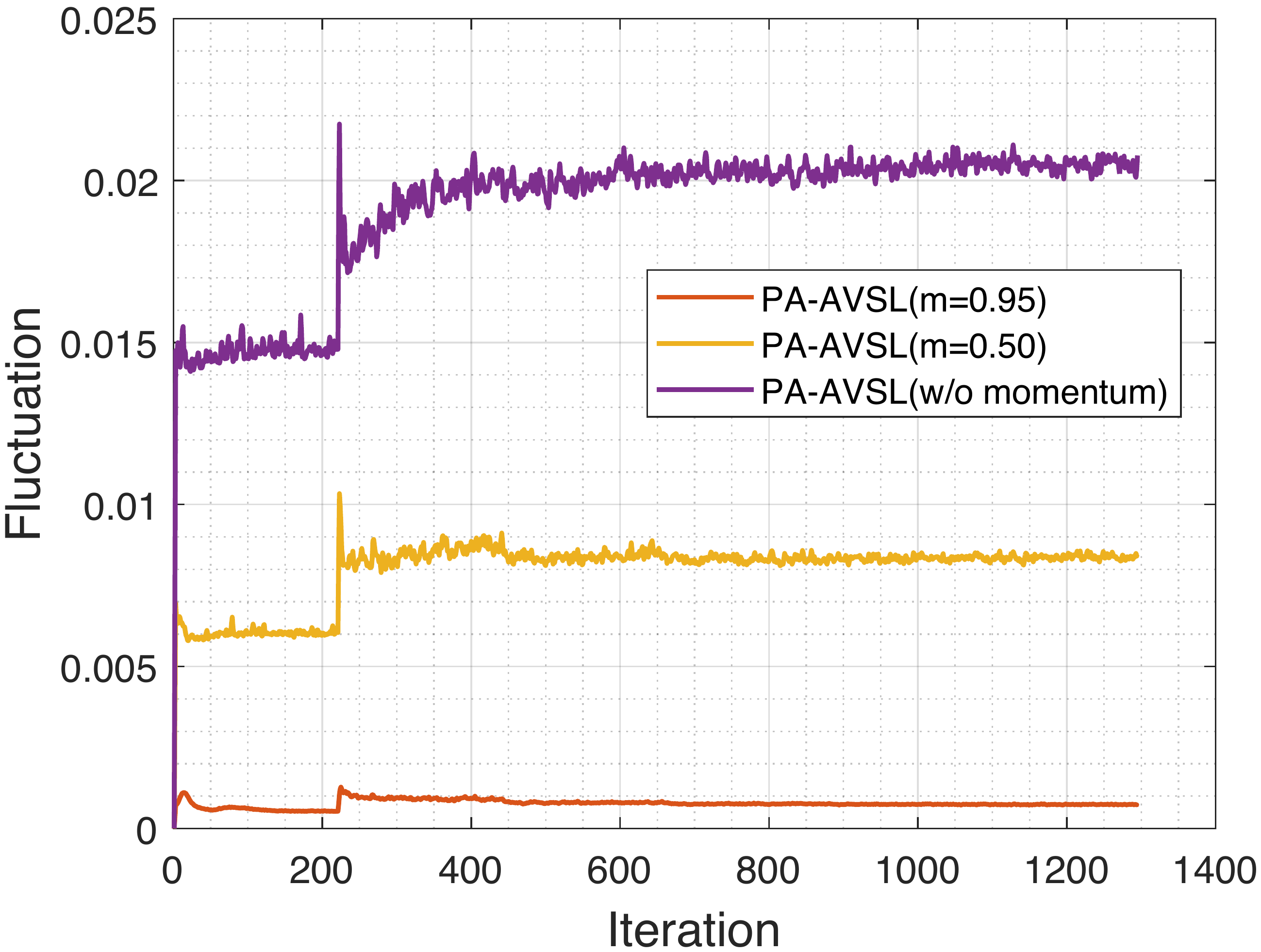}
    \vspace{-3mm}
    \caption{Fluctuation of edges during training.}
    \label{fluctuation}
\end{figure}

\begin{table}[t]
    \centering
    \vspace{-3mm}
    \caption{Ablation study of the edge construction.}
    \vspace{-3mm}
    \label{tab: edge}
    \begin{tabular}{lcccc}
        \toprule
        Method & R@1 & R@2 & R@4  & R@8 \\
        \midrule 
        PA & 87.7 & 92.9 & 95.8 & 97.9 \\
        PA-AVSL (w/o MUS) & 91.0 & 94.6 & 96.7 & 98.1 \\
        PA-AVSL ($\gamma=0.50$) & 91.5 & 95.0 & 97.0 & \textbf{98.4} \\
        \rowcolor{gray!20}
        PA-AVSL ($\gamma=0.95$) & \textbf{91.6} & \textbf{95.2} & \textbf{97.2} & \textbf{98.4} \\
        \bottomrule
    \end{tabular}
    \vspace{-5mm}
\end{table}

\subsection{Detailed Visualization}

We provide more detailed visualization results of the similarity inferring and attribution.
we randomly selected two sample pairs from CUB-200-2011~\cite{wah2011caltech} and Cars196~\cite{krause20133d} for similarity attribution, as shown in Figure~\ref{fig:visualization}.
From top to bottom, we first selected the top-128 reliable nodes with a high $p^l_i$ among all the 512 nodes and further displayed the three most similar nodes framed in green dotted boxes and the three most dissimilar ones framed in red dotted boxes. 
Subsequently, we decompose one unreliable node to the adjacent related nodes.
We quantitatively show the reliabilities $p^l_i$, similarity nodes $\delta^l_i$, and sensitivities $\lambda^l_i$ under each box.

We observe that the similarity nodes with higher sensitivity value $\lambda^l_i$ are more likely positioned in higher layers, which correspond to clearer concepts such as "wing", "head", and "feet" as shown on the left of Figure~\ref{fig:visualization}.
In addition, patterns of low-level features are relatively difficult to recognize.
This demonstrates that high-level features tend to encode object-level concepts while low-level features focus on pixel-level concepts.
In addition, we discover that nodes and concepts may not correspond to each other one-to-one.
For example, multiple nodes may all focus on the ``wheel" part of cars as shown on the right of Figure~\ref{fig:visualization},
which indicates that concepts extracted by CNNs are not completely consistent with humans.

\subsection{Further Analysis}

\paragraph{The strategy of edge construction :}

Due to the image noise, computing edges only based on a single sample may cause large fluctuation.
Instead, we propose to learn the edges dependent on the entire dataset.
We adopt a momentum updating strategy (MUS) to filter the image noise formulated by \eqref{eq:momentum}.
We plot the fluctuation amplitude curves of edges in Figure~\ref{fluctuation}
and see that the proposed momentum updating strategy obtains more stable edges.
We further conducted an ablation study (Table~\ref{tab: edge}) to analyze the influence of MUS on the performance, where ``PA-AVSL (w/o MUS)'' denotes our framework without the momentum updating and $\gamma$ is the momentum factor.
We see that using the momentum updating strategy with a large momentum factor leads to the best performance, indicating the importance of stable edges.

\begin{table}[t]
    \centering
    \vspace{-3mm}
    \caption{Ablation study of the reliability estimation.}
    \label{tab: reliability}
    \vspace{-3mm}
    \begin{tabular}{lcccc}
        \toprule
        Method & R@1 & R@2 & R@4  & R@8 \\
        \midrule 
        PA & 87.7 & 92.9 & 95.8 & 97.9 \\
        PA-AVSL (LR) & 90.9 & 94.6 & 96.6 & 98.0 \\
        \rowcolor{gray!20}
        PA-AVSL & \textbf{91.5} & \textbf{95.0} & \textbf{97.0} & \textbf{98.4} \\
        \bottomrule
    \end{tabular}
    \vspace{2mm}
\end{table}

\begin{figure}[t]
    \centering
    \vspace{-3mm}
    \begin{subfigure}{0.48\linewidth}
        \includegraphics[width=1\linewidth]{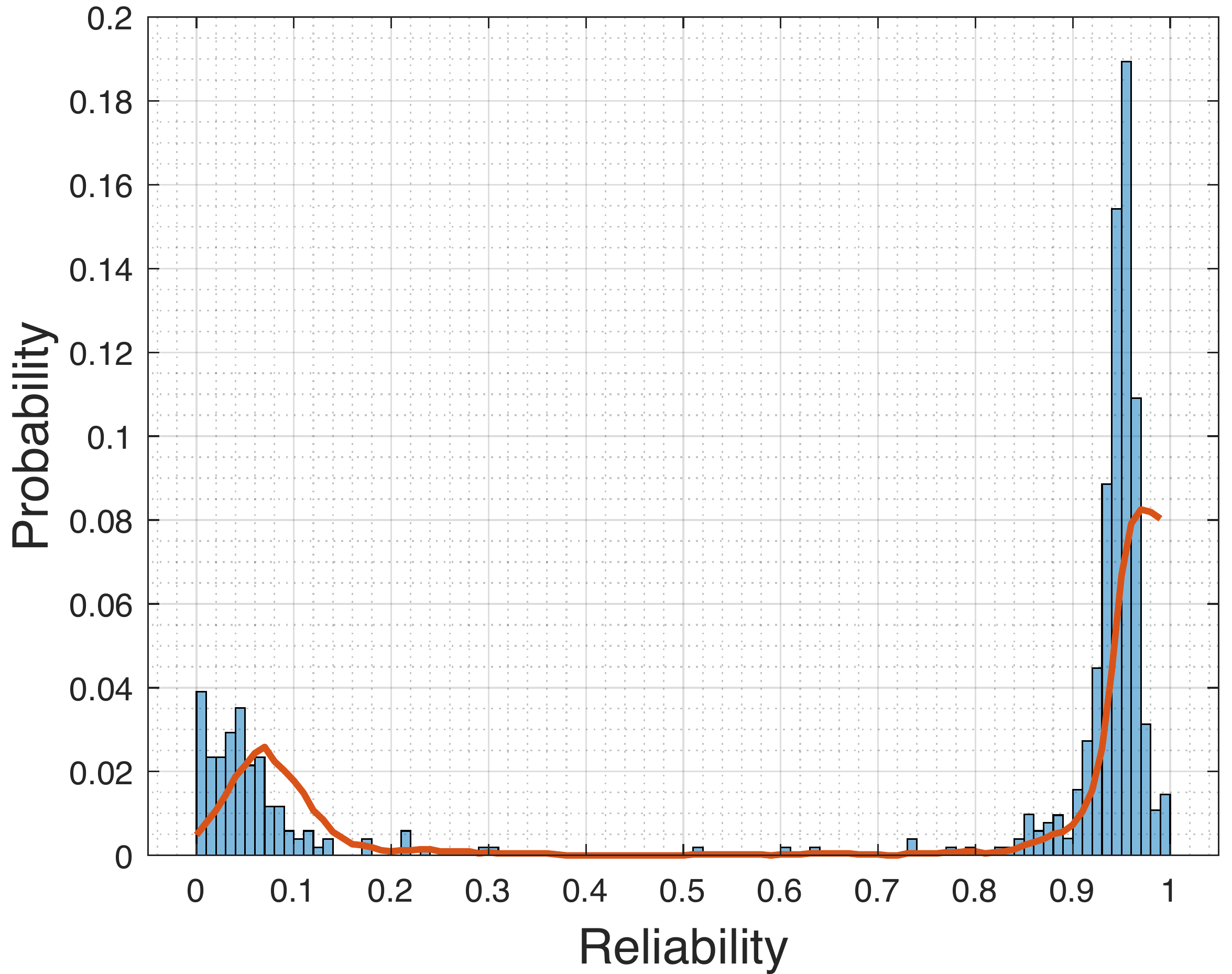}
        \caption{Reliability}
        \label{fig:prob_dist}
    \end{subfigure}
    \begin{subfigure}{0.48\linewidth}
        \includegraphics[width=1\linewidth]{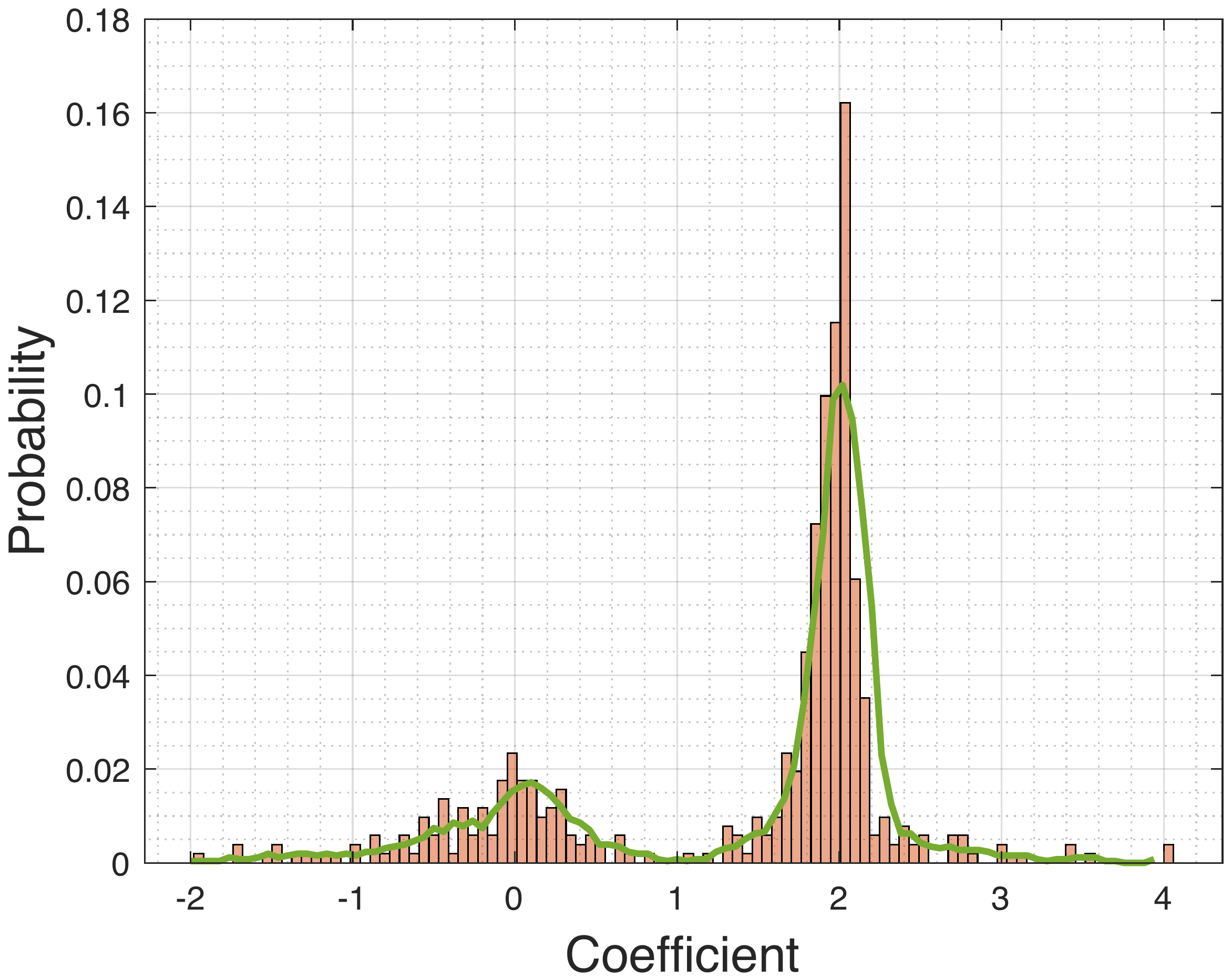}
        \caption{Coefficient}
        \label{fig:coef_dist}
    \end{subfigure}
    \vspace{-4mm}
    \caption{Frequency histogram.}
    \vspace{-5mm}
\end{figure}

\paragraph{The design of reliability:}
We conducted an ablation study about different designs of computing reliability as shown in Table~\ref{tab: reliability},
where ``PA-AVSL (LR)'' represents learning the reliability by a fully-connected layer
(i.e., $\eta_i^l = h(\hat{u}_i^l)h(\hat{u}_i^{'l})$).
The comparison demonstrates that a priori design is more effective than a learning-based one.

\paragraph{The effectiveness of reliability detection:}

We show the distribution of the reliability in Figure~\ref{fig:prob_dist} and observe that only a few significantly unreliable nodes will trigger the top-down rectification.
We further show the distribution of the coefficient of the sigmoid regression as \eqref{eq:reliability} in Figure~\ref{fig:coef_dist}.
The sigmoid regression acts like a filter and assigns a small number of inaccurate reliability estimations with small coefficients close to zero.

{\small
\bibliographystyle{ieee_fullname}

\begin{thebibliography}{10}\itemsep=-1pt

\bibitem{ahonen2006face}
Timo Ahonen, Abdenour Hadid, and Matti Pietikainen.
\newblock Face description with local binary patterns: Application to face
  recognition.
\newblock {\em TPAMI}, 28(12):2037--2041, 2006.

\bibitem{chen2019hybrid}
Binghui Chen and Weihong Deng.
\newblock Hybrid-attention based decoupled metric learning for zero-shot image
  retrieval.
\newblock In {\em CVPR}, pages 2750--2759, 2019.

\bibitem{chen2020adapting}
Lei Chen, Jianhui Chen, Hossein Hajimirsadeghi, and Greg Mori.
\newblock Adapting grad-cam for embedding networks.
\newblock In {\em WACV}, pages 2794--2803, 2020.

\bibitem{chen2017beyond}
Weihua Chen, Xiaotang Chen, Jianguo Zhang, and Kaiqi Huang.
\newblock Beyond triplet loss: a deep quadruplet network for person
  re-identification.
\newblock In {\em CVPR}, pages 403--412, 2017.

\bibitem{duan2019deep}
Yueqi Duan, Lei Chen, Jiwen Lu, and Jie Zhou.
\newblock Deep embedding learning with discriminative sampling policy.
\newblock In {\em CVPR}, pages 4964--4973, 2019.

\bibitem{duan2018deep}
Yueqi Duan, Wenzhao Zheng, Xudong Lin, Jiwen Lu, and Jie Zhou.
\newblock Deep adversarial metric learning.
\newblock In {\em CVPR}, pages 2780--2789, 2018.

\bibitem{hadsell2006dimensionality}
Raia Hadsell, Sumit Chopra, and Yann LeCun.
\newblock Dimensionality reduction by learning an invariant mapping.
\newblock In {\em CVPR}, volume~2, pages 1735--1742, 2006.

\bibitem{harwood2017smart}
Ben Harwood, Vijay Kumar~BG, Gustavo Carneiro, Ian Reid, and Tom Drummond.
\newblock Smart mining for deep metric learning.
\newblock In {\em ICCV}, pages 2821--2829, 2017.

\bibitem{he2016deep}
Kaiming He, Xiangyu Zhang, Shaoqing Ren, and Jian Sun.
\newblock Deep residual learning for image recognition.
\newblock In {\em CVPR}, pages 770--778, 2016.

\bibitem{hermans2017defense}
Alexander Hermans, Lucas Beyer, and Bastian Leibe.
\newblock In defense of the triplet loss for person re-identification.
\newblock {\em arXiv}, abs/1703.07737, 2017.

\bibitem{hu2014discriminative}
Junlin Hu, Jiwen Lu, and Yap-Peng Tan.
\newblock Discriminative deep metric learning for face verification in the
  wild.
\newblock In {\em CVPR}, pages 1875--1882, 2014.

\bibitem{huang2016local}
Chen Huang, Chen~Change Loy, and Xiaoou Tang.
\newblock Local similarity-aware deep feature embedding.
\newblock {\em arXiv}, abs/1610.08904, 2016.

\bibitem{kim2020proxy}
Sungyeon Kim, Dongwon Kim, Minsu Cho, and Suha Kwak.
\newblock Proxy anchor loss for deep metric learning.
\newblock In {\em CVPR}, pages 3238--3247, 2020.

\bibitem{kim2018attention}
Wonsik Kim, Bhavya Goyal, Kunal Chawla, Jungmin Lee, and Keunjoo Kwon.
\newblock Attention-based ensemble for deep metric learning.
\newblock In {\em ECCV}, pages 736--751, 2018.

\bibitem{krause20133d}
Jonathan Krause, Michael Stark, Jia Deng, and Li Fei-Fei.
\newblock 3d object representations for fine-grained categorization.
\newblock In {\em ICCVW}, pages 554--561, 2013.

\bibitem{krizhevsky2012imagenet}
Alex Krizhevsky, Ilya Sutskever, and Geoffrey~E Hinton.
\newblock Imagenet classification with deep convolutional neural networks.
\newblock In {\em NeurIPS}, pages 1097--1105, 2012.

\bibitem{lake2015human}
Brenden~M Lake, Ruslan Salakhutdinov, and Joshua~B Tenenbaum.
\newblock Human-level concept learning through probabilistic program induction.
\newblock {\em Science}, 350(6266):1332--1338, 2015.

\bibitem{lin2017consistent}
Ji Lin, Liangliang Ren, Jiwen Lu, Jianjiang Feng, and Jie Zhou.
\newblock Consistent-aware deep learning for person re-identification in a
  camera network.
\newblock In {\em CVPR}, pages 5771--5780, 2017.

\bibitem{lin2018deep}
Xudong Lin, Yueqi Duan, Qiyuan Dong, Jiwen Lu, and Jie Zhou.
\newblock Deep variational metric learning.
\newblock In {\em ECCV}, pages 689--704, 2018.

\bibitem{loshchilov2017decoupled}
Ilya Loshchilov and Frank Hutter.
\newblock Decoupled weight decay regularization.
\newblock {\em arXiv}, abs/1711.05101, 2017.

\bibitem{lowe2004distinctive}
David~G Lowe.
\newblock Distinctive image features from scale-invariant keypoints.
\newblock {\em IJCV}, 60(2):91--110, 2004.

\bibitem{lu2015multi}
Jiwen Lu, Gang Wang, Weihong Deng, Pierre Moulin, and Jie Zhou.
\newblock Multi-manifold deep metric learning for image set classification.
\newblock In {\em CVPR}, pages 1137--1145, 2015.

\bibitem{movshovitz2017no}
Yair Movshovitz-Attias, Alexander Toshev, Thomas~K Leung, Sergey Ioffe, and
  Saurabh Singh.
\newblock No fuss distance metric learning using proxies.
\newblock In {\em ICCV}, pages 360--368, 2017.

\bibitem{oh2016deep}
Hyun Oh~Song, Yu Xiang, Stefanie Jegelka, and Silvio Savarese.
\newblock Deep metric learning via lifted structured feature embedding.
\newblock In {\em CVPR}, pages 4004--4012, 2016.

\bibitem{opitz2018deep}
Michael Opitz, Georg Waltner, Horst Possegger, and Horst Bischof.
\newblock Deep metric learning with bier: Boosting independent embeddings
  robustly.
\newblock {\em TPAMI}, 42(2):276--290, 2018.

\bibitem{paszke2019pytorch}
Adam Paszke, Sam Gross, Francisco Massa, Adam Lerer, James Bradbury, Gregory
  Chanan, Trevor Killeen, Zeming Lin, Natalia Gimelshein, Luca Antiga, et~al.
\newblock Pytorch: An imperative style, high-performance deep learning library.
\newblock {\em arXiv}, abs/1912.01703, 2019.

\bibitem{qian2019softtriple}
Qi Qian, Lei Shang, Baigui Sun, Juhua Hu, Hao Li, and Rong Jin.
\newblock Softtriple loss: Deep metric learning without triplet sampling.
\newblock In {\em ICCV}, pages 6450--6458, 2019.

\bibitem{ribeiro2016should}
Marco~Tulio Ribeiro, Sameer Singh, and Carlos Guestrin.
\newblock " why should i trust you?" explaining the predictions of any
  classifier.
\newblock In {\em KDD}, pages 1135--1144, 2016.

\bibitem{russakovsky2015imagenet}
Olga Russakovsky, Jia Deng, Hao Su, Jonathan Krause, Sanjeev Satheesh, Sean Ma,
  Zhiheng Huang, Andrej Karpathy, Aditya Khosla, Michael Bernstein, et~al.
\newblock Imagenet large scale visual recognition challenge.
\newblock {\em IJCV}, 115(3):211--252, 2015.

\bibitem{schroff2015facenet}
Florian Schroff, Dmitry Kalenichenko, and James Philbin.
\newblock Facenet: A unified embedding for face recognition and clustering.
\newblock In {\em CVPR}, pages 815--823, 2015.

\bibitem{selvaraju2017grad}
Ramprasaath~R Selvaraju, Michael Cogswell, Abhishek Das, Ramakrishna Vedantam,
  Devi Parikh, and Dhruv Batra.
\newblock Grad-cam: Visual explanations from deep networks via gradient-based
  localization.
\newblock In {\em ICCV}, pages 618--626, 2017.

\bibitem{sohn2016improved}
Kihyuk Sohn.
\newblock Improved deep metric learning with multi-class n-pair loss objective.
\newblock In {\em NeurIPS}, pages 1857--1865, 2016.

\bibitem{stylianou2019visualizing}
Abby Stylianou, Richard Souvenir, and Robert Pless.
\newblock Visualizing deep similarity networks.
\newblock In {\em WACV}, pages 2029--2037, 2019.

\bibitem{sun2020circle}
Yifan Sun, Changmao Cheng, Yuhan Zhang, Chi Zhang, Liang Zheng, Zhongdao Wang,
  and Yichen Wei.
\newblock Circle loss: A unified perspective of pair similarity optimization.
\newblock In {\em CVPR}, pages 6398--6407, 2020.

\bibitem{taigman2014deepface}
Yaniv Taigman, Ming Yang, Marc'Aurelio Ranzato, and Lior Wolf.
\newblock Deepface: Closing the gap to human-level performance in face
  verification.
\newblock In {\em CVPR}, pages 1701--1708, 2014.

\bibitem{verma2012learning}
Nakul Verma, Dhruv Mahajan, Sundararajan Sellamanickam, and Vinod Nair.
\newblock Learning hierarchical similarity metrics.
\newblock In {\em CVPR}, pages 2280--2287, 2012.

\bibitem{wah2011caltech}
Catherine Wah, Steve Branson, Peter Welinder, Pietro Perona, and Serge~J
  Belongie.
\newblock The {Caltech-UCSD Birds-200-2011} dataset.
\newblock Technical Report CNS-TR-2011-001, California Institute of Technology,
  2011.

\bibitem{wan2020nbdt}
Alvin Wan, Lisa Dunlap, Daniel Ho, Jihan Yin, Scott Lee, Henry Jin, Suzanne
  Petryk, Sarah~Adel Bargal, and Joseph~E Gonzalez.
\newblock Nbdt: neural-backed decision trees.
\newblock {\em arXiv}, abs/2004.00221, 2020.

\bibitem{wang2017deep}
Jian Wang, Feng Zhou, Shilei Wen, Xiao Liu, and Yuanqing Lin.
\newblock Deep metric learning with angular loss.
\newblock In {\em ICCV}, pages 2593--2601, 2017.

\bibitem{wang2019multi}
Xun Wang, Xintong Han, Weilin Huang, Dengke Dong, and Matthew~R Scott.
\newblock Multi-similarity loss with general pair weighting for deep metric
  learning.
\newblock In {\em CVPR}, pages 5022--5030, 2019.

\bibitem{wu2017sampling}
Chao-Yuan Wu, R Manmatha, Alexander~J Smola, and Philipp Krahenbuhl.
\newblock Sampling matters in deep embedding learning.
\newblock In {\em ICCV}, pages 2840--2848, 2017.

\bibitem{wu2018beyond}
Mike Wu, Michael Hughes, Sonali Parbhoo, Maurizio Zazzi, Volker Roth, and
  Finale Doshi-Velez.
\newblock Beyond sparsity: Tree regularization of deep models for
  interpretability.
\newblock In {\em AAAI}, volume~32, 2018.

\bibitem{xu2019deep}
Xinyi Xu, Yanhua Yang, Cheng Deng, and Feng Zheng.
\newblock Deep asymmetric metric learning via rich relationship mining.
\newblock In {\em CVPR}, pages 4076--4085, 2019.

\bibitem{ye2016makes}
Han-Jia Ye, De-Chuan Zhan, Xue-Min Si, Yuan Jiang, and Zhi-Hua Zhou.
\newblock What makes objects similar: A unified multi-metric learning approach.
\newblock In {\em NeurIPS}, pages 1235--1243, 2016.

\bibitem{yosinski2014transferable}
Jason Yosinski, Jeff Clune, Yoshua Bengio, and Hod Lipson.
\newblock How transferable are features in deep neural networks?
\newblock {\em arXiv}, abs/1411.1792, 2014.

\bibitem{yosinski2015understanding}
Jason Yosinski, Jeff Clune, Anh Nguyen, Thomas Fuchs, and Hod Lipson.
\newblock Understanding neural networks through deep visualization.
\newblock {\em arXiv}, abs/1506.06579, 2015.

\bibitem{yu2018hard}
Rui Yu, Zhiyong Dou, Song Bai, Zhaoxiang Zhang, Yongchao Xu, and Xiang Bai.
\newblock Hard-aware point-to-set deep metric for person re-identification.
\newblock In {\em ECCV}, pages 188--204, 2018.

\bibitem{yuan2019signal}
Tongtong Yuan, Weihong Deng, Jian Tang, Yinan Tang, and Binghui Chen.
\newblock Signal-to-noise ratio: A robust distance metric for deep metric
  learning.
\newblock In {\em CVPR}, pages 4815--4824, 2019.

\bibitem{yuan2017hard}
Yuhui Yuan, Kuiyuan Yang, and Chao Zhang.
\newblock Hard-aware deeply cascaded embedding.
\newblock In {\em ICCV}, pages 814--823, 2017.

\bibitem{zeiler2014visualizing}
Matthew~D Zeiler and Rob Fergus.
\newblock Visualizing and understanding convolutional networks.
\newblock In {\em ECCV}, pages 818--833, 2014.

\bibitem{zeiler2010deconvolutional}
Matthew~D Zeiler, Dilip Krishnan, Graham~W Taylor, and Rob Fergus.
\newblock Deconvolutional networks.
\newblock In {\em CVPR}, pages 2528--2535, 2010.

\bibitem{zhang2018interpreting}
Quanshi Zhang, Ruiming Cao, Feng Shi, Ying~Nian Wu, and Song-Chun Zhu.
\newblock Interpreting cnn knowledge via an explanatory graph.
\newblock In {\em AAAI}, volume~32, 2018.

\bibitem{zhang2018interpretable}
Quanshi Zhang, Ying~Nian Wu, and Song-Chun Zhu.
\newblock Interpretable convolutional neural networks.
\newblock In {\em CVPR}, pages 8827--8836, 2018.

\bibitem{zhao2021towards}
Wenliang Zhao, Yongming Rao, Ziyi Wang, Jiwen Lu, and Jie Zhou.
\newblock Towards interpretable deep metric learning with structural matching.
\newblock In {\em ICCV}, pages 9887--9896, 2021.

\bibitem{zheng2019hardness}
Wenzhao Zheng, Zhaodong Chen, Jiwen Lu, and Jie Zhou.
\newblock Hardness-aware deep metric learning.
\newblock In {\em CVPR}, pages 72--81, 2019.

\bibitem{zheng2021deep1}
Wenzhao Zheng, Chengkun Wang, Jiwen Lu, and Jie Zhou.
\newblock Deep compositional metric learning.
\newblock In {\em CVPR}, pages 9320--9329, 2021.

\bibitem{zheng2021deep2}
Wenzhao Zheng, Borui Zhang, Jiwen Lu, and Jie Zhou.
\newblock Deep relational metric learning.
\newblock In {\em ICCV}, pages 12065--12074, 2021.

\bibitem{zhou2016learning}
Bolei Zhou, Aditya Khosla, Agata Lapedriza, Aude Oliva, and Antonio Torralba.
\newblock Learning deep features for discriminative localization.
\newblock In {\em CVPR}, pages 2921--2929, 2016.

\bibitem{zhu2019visual}
Sijie Zhu, Taojiannan Yang, and Chen Chen.
\newblock Visual explanation for deep metric learning.
\newblock {\em arXiv}, abs/1909.12977, 2019.

\end{thebibliography}

}

\end{document}